\documentclass[journal,twocolumn]{IEEEtran}

\usepackage{graphicx}
\usepackage{amsmath}
\usepackage{subfigure}
\usepackage{caption}
\usepackage{algorithm}
\usepackage{algpseudocode}
\usepackage{multirow}
\usepackage{setspace}
\usepackage{color,hyperref}

\ifCLASSINFOpdf
\else
\fi

\definecolor{darkblue}{rgb}{0.0,0.0,1.0}
\hypersetup{colorlinks,breaklinks,
	linkcolor=darkblue,urlcolor=darkblue,
	anchorcolor=darkblue,citecolor=darkblue}

\begin{document}

\title{Tracking as A Whole: Multi-Target Tracking by Modeling Group Behavior with Sequential Detection}

\author {  % stops a space
		Yuan~Yuan,~\IEEEmembership{Senior Member,~IEEE,}
		Yuwei~Lu,
		Qi~Wang$^*$,~\IEEEmembership{Senior Member,~IEEE}	     
\IEEEcompsocitemizethanks{\IEEEcompsocthanksitem This work was supported in part by the National Natural Science Foundation of China under Grant 61379094.
	
	The authors are with the School of Computer Science and Center for OPTical IMagery Analysis and Learning (OPTIMAL), Northwestern Polytechnical University, Xi'an, China. 

	 Qi Wang is the corresponding author (e-mail:crabwq@nwpu.edu.cn)
	 
	 \copyright 2017 IEEE. Personal use of this material is permitted. Permission from IEEE must be obtained for all other uses, in any current or future media, including reprinting/republishing this material for advertising or promotional purposes, creating new collective works, for resale or redistribution to servers or lists, or reuse of any copyrighted component of this work in other works.
}}

\markboth{{IEEE} Transactions on Intelligent Transportation Systems}%
{Shell \MakeLowercase{\textit{et al.}}: Bare Demo of IEEEtran.cls for Journals}
% make the title area
\maketitle

% As a general rule, do not put math, special symbols or citations
% in the abstract or keywords.
\begin{abstract}

Video-based vehicle detection and tracking is one of the most important components for Intelligent Transportation Systems (ITS). When it comes to road junctions, the problem becomes even more difficult due to the occlusions and complex interactions  among vehicles. In order to get a precise detection and tracking result, in this work we   propose a novel tracking-by-detection framework. In the detection stage, we present a sequential detection model to deal with serious occlusions. In the tracking stage, we model group behavior to  treat complex interactions with overlaps and ambiguities. The main contributions of this paper are twofold: 1) Shape prior is exploited in the sequential detection model to tackle occlusions in crowded scene. 2)  Traffic force is defined in the traffic scene to model group behavior, and it can assist to handle complex interactions among vehicles.  We evaluate the proposed approach on real surveillance videos at road junctions and the performance has demonstrated the effectiveness of our method.

\end{abstract}

% Note that keywords are not normally used for peerreview papers.
\begin{IEEEkeywords}
multi-target, vehicle detection, tracking, road junction, sequential detection, group behavior
\end{IEEEkeywords}

\IEEEpeerreviewmaketitle

\section{Introduction}

Intelligent Transportation System (ITS) will be the development direction of the future traffic system. With the popularity of monitoring equipment, more and more traffic videos and images have to be analyzed. Faced with the large amount of data, traditional manual management has become unrealistic and  automatic analysis is incrementally necessary as a consequence.

Among the various techniques that enable  ITS, detecting and tracking vehicles are fundamentally important. They can help get the information of the primary traffic occupations and infer the quantitative statistics of traffic status. The goal of this work is to automatically detect and track each individual vehicle in the surveillance video of a road junction, where the traffic condition is more complex and achieving an effective control is more essential.

Several challenges render this problem very difficult. First, vehicles have complex dynamics in the field of camera view. Second, occlusion is very serious between vehicles in the real traffic scene. Third, road intersections have much more kinds of objects and complex environment that will lead to ambiguities during detecting vehicles.   Under these difficulties, the first important thing is to detect most appeared   vehicles and  then track their movements. Thus an excellent method is supposed to detect and track targets as many as possible. However, traditional trackers, such as \cite{particlefilter,WangFY14tracking},  either ignore detection  or  many of them \cite{pamiMeiL11sparse,prBaiL12,YuanFW14super} annotate the target by hand in the first frame of video sequence. These manual methods are impractical for real traffic videos, because too many targets   exit in the field of view and new ones keep emerging. As a result, discriminative tracking methods with online learning \cite{ensemble,prJangCTTK15} are proposed. In such approaches, a  specific detector is trained in a semi-supervised fashion and then used to find out the object in continuous frames. However, these algorithms only focus on single target without considering the multi-target situation. Several techniques \cite{MilanRS14pami,7353188,multrackingnetworkflows,7448472} are consequently dished to deal with multi-target tracking by optimizing detection assignments over a temporal window. Such approaches apply off-line trained detectors to locate the targets and associate them with their tracks. Although they can overcome several difficulties such as the uncertainty in the number of targets and template drift, they are still inadequate when facing   occlusion.          Particularly, when tracking a crowd of vehicles in the traffic surveillance video, the data association often fails in the aforementioned methods because of partial occlusions and complex interactions with overlaps and ambiguities.    Similar to our approach, some methods use Social Force Model \cite{09sfm} to improve tracking results, e.g. \cite{cvprQinS12,iccvSFM09}. Our approach is different than \cite{cvprQinS12,iccvSFM09} in that we model group behavior based on traffic force. The difference between them will be discussed in \ref{gbm}.

In this work, we deal with such difficulties by proposing a sequential detection model, which explores shape prior segmentation, and integrates tracking with group behavior context. While the deformable part-based model (DPM) \cite{DPMpami} has outstanding performance in VOC challenges \cite{pascalvoc2007}, yet it still has poor performance in crowded scenes. Since there are more complex background and  targets in actual environment, it is hard  to detect all the targets only with one detector. Our sequential detection model utilizes the deformable part-based model whose threshold value ($\eta$) is much smaller as a sieve to obtain more candidates of targets. Meanwhile,   to decrease the inevitable false detections, a shape prior based segmentation is put forward  to refine the   results of DPM.  
% We call this  procedure as generalized DPM (GDPM).

\begin{figure*}
	\includegraphics[width=1\textwidth,height=6.5cm]{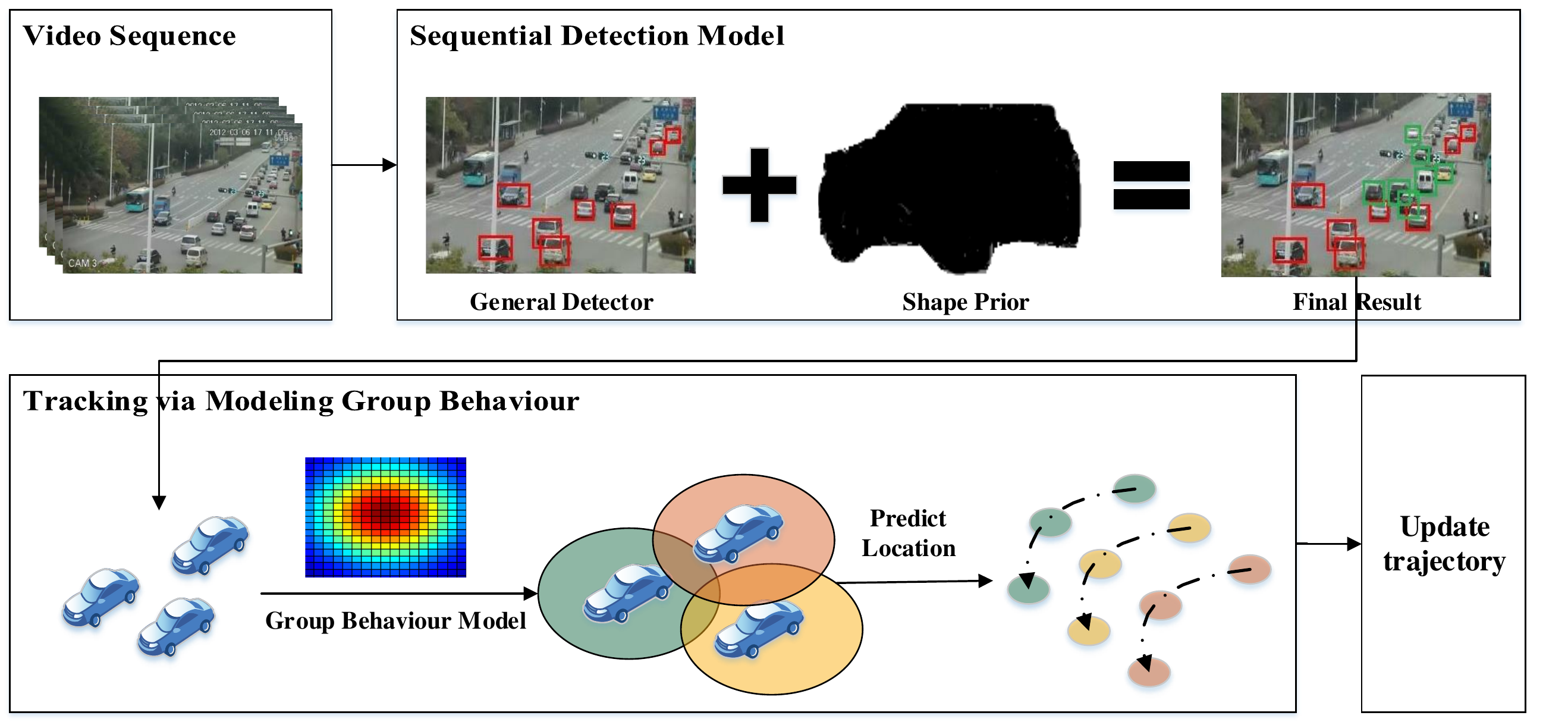}
	\caption{\label{framework} Overview of the proposed tracking-by-detection framework.}
\end{figure*}

Due to the lower threshold DPM,   the partial occlusion can be handled more robustly.  However, complex dynamics  still bothers  us. In the actual environment, each single vehicle is impacted by the others and the speed of one vehicle will be affected by its neighbors. In this case,  traditional trackers which track vehicles without taking their neighbors into account will be more likely to drift because they do not capture the intrinsic properties of the vehicle movements. Therefore, we model group behavior that simultaneously  considers both the individual vehicle and its neighboring context to improve the performance of the tracker.       Especially, our group model acts on vehicles with the same direction.   The group behavior model is based on the distance between two vehicles. The closer two vehicles are, the greater repulsive force between them will be. Since the contextual information is added, the group behavior model (GBM) will assist to predict possible locations of target more reasonably.

The proposed tracking-by-detection framework consists of the steps illustrated in Fig. \ref{framework}. First, a general vehicle  detector is used. The detector is supposed to provide candidates of vehicles as many as possible   on each frame, no matter whether including false detections or not. In this work, we utilize DPM as an example to implement vehicle detection. Second, the shape prior information is employed to refine the detection results.  An energy function is formulated in the graph-cut segmentation by incorporating the vehicle shape prior. Through  minimizing the energy function, vehicles can be distinguished from false detections.  After that, a GBM-based tracker contributes to predicting locations of the detected vehicles with taking the influences of the neighboring vehicles into account. Owing to modeling group behavior, we can obtain more reasonable vehicle locations. Finally, these predicted locations will be assigned to trajectories and  the tracked vehicles will be updated.

In summary, faced with the actual scene of road junctions, we extend traditional approaches so as to adapt to a crowded scene. The main contributions of this paper are as follows. First, we adopt a novel sequential detection model which exploits a shape prior segmentation to tackle occlusion in crowded scene. Second, the complex group behavior in traffic scene is modeled by traffic force to handle influence of the neighboring vehicles.

The rest of this paper is organized as follows. Section \ref{related} introduces the related work while  Section \ref{our_method} describes the proposed approach. Experimental results are discussed in Section \ref{experiment} and conclusion and future direction are presented in Section \ref{conclusion}.

\section{Related Work} \label{related}
Object detection and tracking have a long history in computer vision. Much progress has been  made in recent years. Since this work is mainly about multi-target detection and tracking, we review existing works in terms of two main categories: vehicle detection and multi-target tracking.

\subsection{Vehicle Detection}

Object detection has a very wide range of applications. In this paper, we mainly discuss vehicles in the traffic scene, so we will just review the vehicle detection methods instead of general object detection. Vision-based vehicle detection for traffic surveillance video has received considerable attention. As a rigid target, vehicles have significant structural characteristics, which is more stable than flexible objects. In this paper, we follow the common  two steps in vehicle detection  \cite{review}: Hypothesis Generation (HG) and Hypothesis Verification (HV).

\subsubsection{Hypothesis Generation}

The goal of   HG step is to find candidate vehicle locations in an image quickly for further exploration. HG approaches can be mainly classified into the following three types: knowledge-based, stereo-based and motion-based.

%Knowledge-based methods make use of a priori knowledge to hypothesize target locations. Different kinds of priori information are used to identify vehicles. Bertozzi \textit{et al.} \cite{bertozziSymmetry} study symmetry. However, symmetry estimations are sensitive to noise in homogeneous areas. Therefore, color cue is applied in \cite{Guocolor} to extract vehicles from background. Apart from color features, shadow information is also employed in \cite{shadowMori} to hypothesize vehicle locations. Inevitably, the intensity of the shadow is influenced by light conditions. Shadow prior is destined not to have an excellent performance for vehicle detection. Considering the fact that vehicles always have a fixed shape, the structural information is utilized in a large amount. Massimo \textit{et al.} \cite{BertozziCorner} propose a corner-based method which makes use of four corners of the vehicle for vehicle detection. Betke \textit{et al.} \cite{BetkeHD00edgeinfo} use edge detection to find vertical and horizontal structures of the vehicle. Moreover, Thomas \textit{et al.} \cite{Kalinke98atexturebased} employ texture information as a cue to narrow down the search area for detecting vehicles.

Knowledge-based methods make use of a priori knowledge to hypothesize target locations. Different kinds of priori information are used to identify vehicles. Teoh \textit{et al.} \cite{bertozziSymmetry} study symmetry. However, symmetry estimations are sensitive to noise in homogeneous areas. Therefore, shadow information is applied in \cite{shadowMori} to hypothesize vehicle locations. Inevitably, the intensity of the shadow is influenced by light conditions. Shadow prior is destined not to have an excellent performance for vehicle detection. Apart from shadow, the structural information is utilized in a large amount. Wu \textit{et al.} \cite{BetkeHD00edgeinfo} use edge detection to find moving vehicles on the road. Moreover, color feature becomes popular after aggregate channel features (ACF) are proposed in \cite{DollarPAMI14pyramids}. The ACF proposes 10 channel features including 3 LUV color channels, and this feature is outstanding in vehicle detection.     Ohn-Bar \textit{et al.} \cite{ohnbar14} employ ACF and clustering method to narrow down  the search area for detecting vehicles.

%color cue is applied in \cite{Guocolor} to extract vehicles from background. Apart from color features, shadow information is also employed in \cite{shadowMori} to hypothesize vehicle locations. Inevitably, the intensity of the shadow is influenced by light conditions. Shadow prior is destined not to have an excellent performance for vehicle detection. Considering the fact that vehicles always have a fixed shape, the structural information is utilized in a large amount. Massimo \textit{et al.} \cite{BertozziCorner} propose a corner-based method which makes use of four corners of the vehicle for vehicle detection. Betke \textit{et al.} \cite{BetkeHD00edgeinfo} use edge detection to find vertical and horizontal structures of the vehicle. Moreover, Thomas \textit{et al.} \cite{Kalinke98atexturebased} employ texture information as a cue to narrow down the search area for detecting vehicles.    Massimo \textit{et al.} \cite{BertozziCorner} propose a corner-based method which makes use of four corners of the vehicle for vehicle detection.

%ohnbar14 DollarPAMI14pyramids

Stereo vision-based methods \cite{MandelbaumMBRH98,BertozziBIPM,WangFY15} apply stereo information for vehicle detection in two ways. One is disparity map, while the other is Inverse Perspective Mapping (IPM), an anti-perspective transformation. Lefebvre \textit{et al.} \cite{MandelbaumMBRH98} convert the disparity map into a 3D map to extract 3D points, while Bertozzi \textit{et al.} \cite{BertozziBIPM} employ IPM to acquire stereo vision. Both the disparity and IPM are employed to get contours of targets. With the help of contour information, vehicles can be detected on images.

Motion-based methods employ movement information of vehicles to distinguish vehicles from background. Normally, velocity is the most useful cue to take motion into consideration. Since   vehicles keep moving and background is always static, objects  and background can  be separated according to the difference of  velocity. Martinez \textit{et al.} \cite{KrugerMotionbased} use  optical flow method to estimate velocity of each pixel in the image.  Afterwards, pixels which have the similar velocity will be clustered together, and these clusters of pixels is the hypothesis of vehicles. However, generating a displacement vector for each pixel is time consuming. In contrast to pixel-based optical flow, feature based methods, such as color features \cite{Ritter95motion}, extract features from an image. And then   optical flow of  feature points will be computed. By clustering optical flow of feature points like pixels-based methods,  vehicle hypotheses are generated.  Since not all the velocities of pixels need to be estimated, this makes feature based methods faster than pixel-based ones.

\subsubsection{Hypothesis Verification}

Compared with HG step, the input of  HV step is the set of hypothesized locations from  HG step. For this procedure, tests are performed to verify the correctness of a hypothesis. There are mainly two types of HV approaches: template-based and appearance-based.

%Template-based approaches apply predefined patterns from the vehicle class and perform correlation between the image and the template.  Parodi and Piccioli \cite{Parodi95Feature} propose a template-based method with the presence of license and rear windows. By searching license and rear windows, vehicles will be detected. Leon \textit{et al.} \cite{LeonH12} put forward a template-based approach using mixture of deformable parts models. They expand the original deformable part models (DPM) \cite{DPMpami} to adapt to crowded scenes.

Template-based approaches apply predefined patterns from the vehicle class and perform correlation between the image and the template. Li \textit{et al.} \cite{eccvLiWZ14} propose an And-Or model that integrates context and occlusion for verifying hypotheses. Felzenszwalb \textit{et al.} \cite{DPMpami} propose deformable part models (DPM) to structure template model. Each model is composed of different parts of the object. The system detects objects by scoring each hypothesis according to the similarity between hypothesis and DPM model and thresholding scores. Leon \textit{et al.} \cite{LeonH12} put forward a template-based approach using mixture of deformable parts models. They expand the original DPM \cite{DPMpami} to adapt to crowded scenes. Wang \textit{et al.} \cite{7287759} also propose a probabilistic inference framework based on part models for improving detection performance.

Appearance-based approaches learn the features of the vehicles from a set of training images which should capture the variability in vehicle appearance. Usually, appearance models treat a two-class pattern classification problem: vehicle and nonvehicle. Wu and Zhang \cite{Zhangpca} apply standard Principal Components Analysis (PCA) for extracting global features to detect  vehicles. Owing to small training data set, it is difficult to draw any meaningful conclusions. Li \textit{et al.} \cite{LiYMKG04} employ segmentation and neural network classifier for distinguishing vehicles from background. Khammari \textit{et al.} \cite{Khammari2005Vehicle} add depth image to set up their appearance models. Apart from the observed features, Zheng \textit{et al.} \cite{ZhengL09Strip} design image strip features based on the vehicle structure for vehicle detection. Since features come from the side view of the vehicle, their detector is sensitive to the viewpoint.

\subsection{Multi-Target Tracking}
A significant amount of work has been reported for multi-target tracking. There are two main representative approaches,  detection-based data association and energy minimization.
\subsubsection{Detection-based Data Association}
Detection-based data association regards multi-target tracking as a data association problem. Longer tracklets can be formed by detections between two continuous frames. The most classic framework of this approach is proposed by Nevatia \textit{et al.} \cite{HuangWN08}. They handle data association in three levels. In the low-level, they connect detection responses in continuous frames into short tracklets. A threshold value would be used to exclude unreliable ones. In the mid-level, they compute an affinity score for each reliable tracklet obtained from low-level and connect short tracklets into longer tracklets. In the high-level, a scene structure model is estimated based on the tracklets provided by the middle level. Afterward, with the help of scene knowledge, the long-range trajectory association is performed.

Some work follows this basic framework. Zhang \textit{et al.} \cite{multrackingnetworkflows} define data association as a maximum-a-posteriori problem given a set of object detection results as input observations, while Brendel \textit{et al.} \cite{miltrackingweightset} formulate data association problem as finding the maximum-weight independent set of graph that is built by pairs of detection responses from every two consecutive frames.

\subsubsection{Tracking via Energy Minimization}
Many problems can be transformed into an energy minimization problem. This is true for multi-target tracking. In recent years, several energy minimization-based tracking methods \cite{iccvLeibeSG07,MilanRS14pami} are proposed. In these  methods,  detection responses are known and solution space is the combination of tracklets that are composed of these responses, which is different from common data association methods whose current frame is inferred by previous ones. Milan \textit{et al.} \cite{MilanRS14pami} construct an energy function that  depends on the locations and motions of all targets in all frames for obtaining globally optimal solution  considering physical constraints, such as target dynamics.  By minimizing the energy function, they get the final tracking result.  Leibe \textit{et al.} \cite{iccvLeibeSG07} present a multi-object tracking approach which considers object detection and space-time trajectory estimation as an optimization problem. And the successful trajectory hypotheses are fed back to guide detection in the future frames. Since minimizing energy function is time-consuming, energy minimization-based methods for multi-object tracking  suffer from low computational efficiency.

\section{Our Approach} \label{our_method}

In this section, we will give a detailed explanation of our tracking-by-detection framework. As mentioned before, in the surveillance video of road intersections, occlusions and complex interactions with overlapping and ambiguities are the main difficulties. Hence, in the detection stage, we present a sequential detection model that explores shape prior segmentation to improve the detector in crowded scene. In the tracking stage, on the other hand, traffic force is proposed  to model group behavior. Interactions between individual vehicles are taken into consideration to tackle nonlinear dynamics in vehicle tracking.

\subsection{Sequential Detection Model}

Though object detection has made much progress, existing detection approaches are still not well tailored to crowded scenes. Our motivation comes from boosting algorithm, in which   the single classifier does not work well but combining several weak classifiers to a cascade classifier  can achieve an outstanding performance. In the same way, a single vehicle detection algorithm cannot find out all the targets. Therefore, we combine several basic algorithms having distinctive superiorities to produce a sequential detector. The Sequential Detection Model consists of two main parts, DPM  and shape prior segmentation.

\subsubsection{DPM}
As shown in Fig. \ref{framework}, a general detector is utilized in our framework to get enough possible locations of targets. The reason why we choose DPM are mainly as follows. First, DPM can easily get shape templates with various viewpoints  and describe the target with abundant information. Second, DPM is well suited for occlusions that are serious in the scene of road intersections.

In DPM, the detection score of a hypothesis, $score(h_{obj})$, is given by the filter score    at the examined location minus a deformation cost that depends on the relative position of each part with respect to the root filter  plus the bias, as is shown in Eq.\ref{score}:
\begin{equation} \label{score}
	score(h_{obj})= \sum_{i=0}^nF_i-\sum_{i=1}^nD_i+b,
\end{equation}
where $b$ is a bias term, $n$ is the number of parts, $D_i$ is the deformation cost of part $i$, $F_i$ is the score of each part and $F_0$ represents the root part.

A score represents the similarity between pre-trained model and a hypothesis. DPM get final detections by thresholding score. However, different from the original DPM,  we set  a  low threshold value instead of self-generated one for DPM in detection procedure.     As a result, reducing threshold can obtain  vehicle  candidates   as many as possible.    Due to the significant low threshold value, the  results of DPM have both vehicle and nonvehicle targets. Though we improve the recall rate of detecting vehicles, more false detections appear  inevitably. In order to exclude the false detections, a shape prior segmentation is further applied.

\subsubsection{Shape Prior Segmentation}

As is implied by the name, the shape prior segmentation takes shape information into consideration so as to exclude the false detections. Each detection window obtained from the DPM is processed independently.

Image segmentation can be regarded as a pixel labeling problem  actually. The label of the pixel depends on whether it is in object or background and this process can be achieved by minimizing the energy function through minimum graph cut. Let $L=\{l_1,l_2,...,l_i,...,l_m\}$ be the label set of pixels,  where $l_i$ is the label of the pixel $i$ in the image. The  pixel is assigned label $l_i=1$ if it belongs to object and $l_i=0$ if it belongs to background. The energy function for the shape prior based graph cut is usually defined as the following equation \cite{WangZR13shape,Jingshape}:
\begin{equation} \label{orig_fun}
	E(L) = R(L)+B(L)+E_{shape},
\end{equation}
where, $R(L)$ is the regional term, $B(L)$ is the boundary term and $E_{shape}$ is the shape prior term. Compared to traditional graph cut based segmentation, the shape prior term is added. The goal of shape prior term is to segment targets with similar shape to the template.

In sequential detection model of our framework, shape prior segmentation is applied to distinguish vehicles from other targets. Our shape prior segmentation is just like a refinement. It can remove those false detections and extract vehicles. We define the energy function of graph cut segmentation with shape prior in the following way:
\begin{equation}  \label{energy}
\begin{aligned}
	E(L)=& \sum_{p{\in}P}D_p(l_p)+\sum_{(i,j){\in}N_p:l_i{\not=}l_j}V_{i,j}(l_i,l_j)     \\
	 &+\sum_{(i,j){\in}N_p:l_i{\not=}l_j}E_{i,j}(l_i,l_j), 
\end{aligned}
\end{equation}
where $P$ is the set of all pixels and $N_p$ is the set of pixels in the neighborhood of $p$. $D_p(l_p)$ is the penalty of assigning label $l_p{\in}L$ to $p$, and $V_{i,j}(l_i,l_j)$ is the penalty of labelling the pair $i, j$ with labels $l_i, l_j{\in}L$, respectively. $E_{i,j}(l_i,l_j)$ represents a pairwise shape constraint term that penalizes the difference between the shape template and the target.

To be specific, the  region term $D_p(l_p)$ is:
\begin{eqnarray}
	D_p(l_p=1) = -logP_r(I_p|obj),   \\
	D_p(l_p=0) = -logP_r(I_p|back),
\end{eqnarray}
where $P_r(I_p|obj)$ and $P_r(I_p|back)$ are the probability distributions that can be learned beforehand for both the object and the background, and $I$ represents the pixel intensity.

And the edge term which punishes those pixels with similar intensities is defined as:
\begin{equation}
	V_{i,j}(l_i,l_j) = \exp(-\frac{(I_i-I_j)^2}{2\alpha^2})\frac{1}{dis(i,j)},
\end{equation}
where $\alpha$ can be seen as camera noise, and $dis(i,j)$ is the Euclidean distance between pixels $i$ and  $j$.

When applying adaptive shape prior, we employ the idea of level-set template, and define shape energy term $E_{i,j}(l_i,l_j)$ in Eq.(\ref{energy}) in the following way:
\begin{equation} \label{endseg}
	E_{i,j}(l_i,l_j)={\phi}(\frac{pos_i+pos_j}{2}),
\end{equation}
where ${\phi}$ is a regular, unsigned distance function whose zero level set corresponds to the shape template. $pos_i$ and $pos_j$ are the locations of pixels. $\phi(pos_i)$ will be zero if $l_i=1$, and $\phi(pos_i)$ will be the shortest distance to the shape boundary if $l_i=0$.   Since they are neighboring pixels, they will be along the contour of shape, when minimizing $E_{i,j}$.

\begin{figure}
	\center
	\includegraphics[width=.45\textwidth,height=4cm]{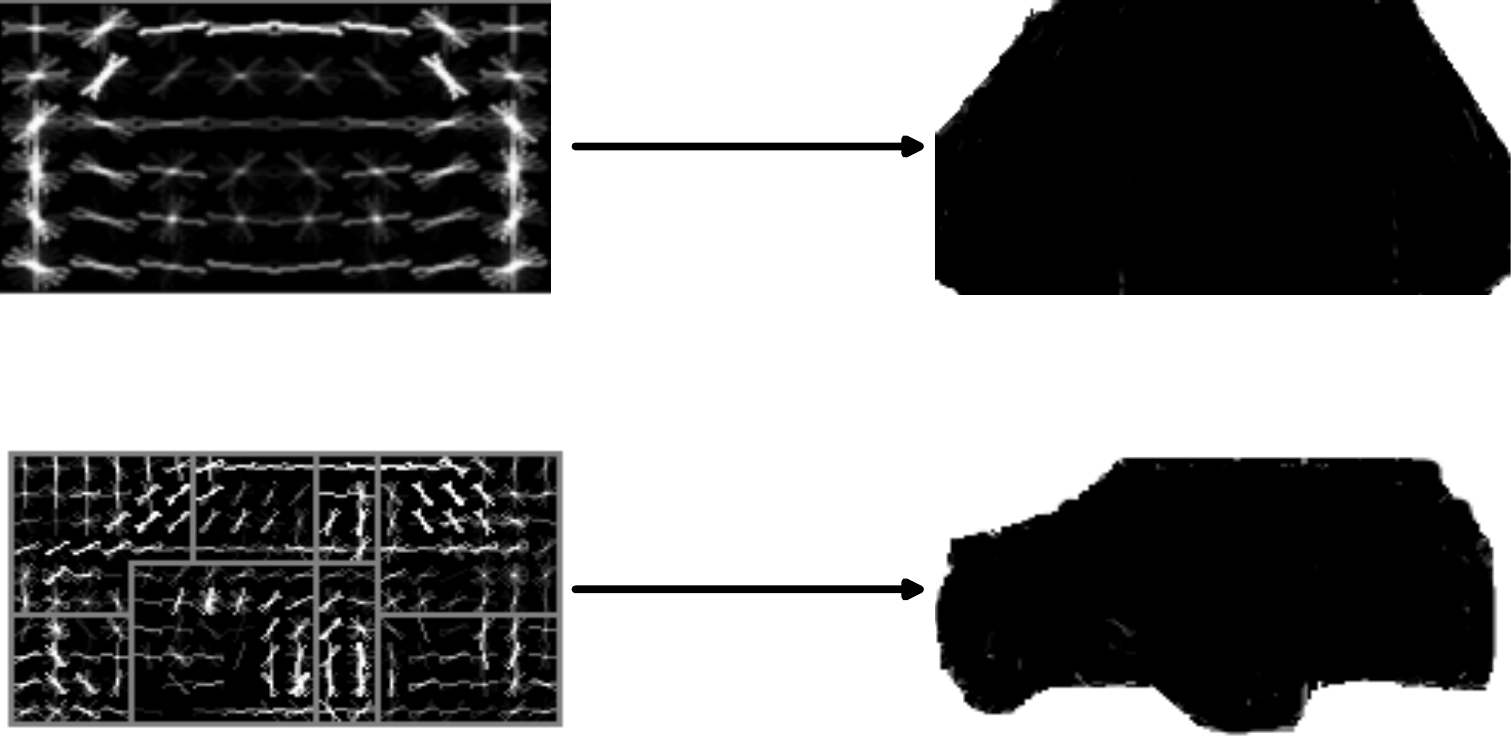}
	\caption{\label{transform}Examplar transformation from HOG template to our binary template.}
	% The left vehicle model is trained by DPM with HOG feature. Our binary template, the right one, is obtained by converting the HOG model.}
\end{figure}

As for the definition of shape template, we employ  binary figures of the vehicles. Since we incorporate shape prior by zero level set, a binary figure will speed up the computation. For this purpose, the shape templates are generally supposed to be trained by samples. Fortunately,  DPM has trained multi-parts from various viewpoints of the vehicle. In this work,  we adopt vehicle models trained by DPM and transform them into binary templates.  These models are trained by VOC-2007 \cite{pascalvoc2007} dataset and our own data.     As a result, when DPM detects a vehicle by one of the parts,  we can utilize the corresponding binary figure   as the shape template. Namely, different shapes of the vehicle are applied depending on the maximum response part used in   the DPM  detection stage. Fig. \ref{transform} shows some examples of our shape templates.

By minimizing the energy function of Eq. \ref{energy}, false detections are removed and vehicles will be detected finally.

%After handled by DPM and shape prior segmentation, we get outstanding performance than original deformable part model.

\subsection{Tracking via Group Behavior Model}
\label{socialcue}
After obtaining the final detections, we employ these detection windows as our observation targets for tracking. As mentioned before, occlusions and complex motions will be the challenges. Owing to part-based model, we can deal with occlusions. However, complex motion model still troubles us.

Under the dynamics, it is hard for us to treat the motion of a vehicle separately. Each vehicle is affected by its surroundings and  regarding the vehicles as a group in the scene of road intersections is more reasonable. In view of the above fact, we attempt to model the group behavior of vehicles to assist tracking procedure. When tracking a vehicle in such a group, we should consider not only the state of the vehicle itself, but also the influence of other vehicles in the group.

\subsubsection{Group Behavior Model}  \label{gbm}
\label{GBMSection}
\begin{figure*}
	\centering
	\subfigure[The state-space model of traditional tracking.]{
		\label{oldtra}
		\includegraphics[width=0.4\textwidth,height=3cm]{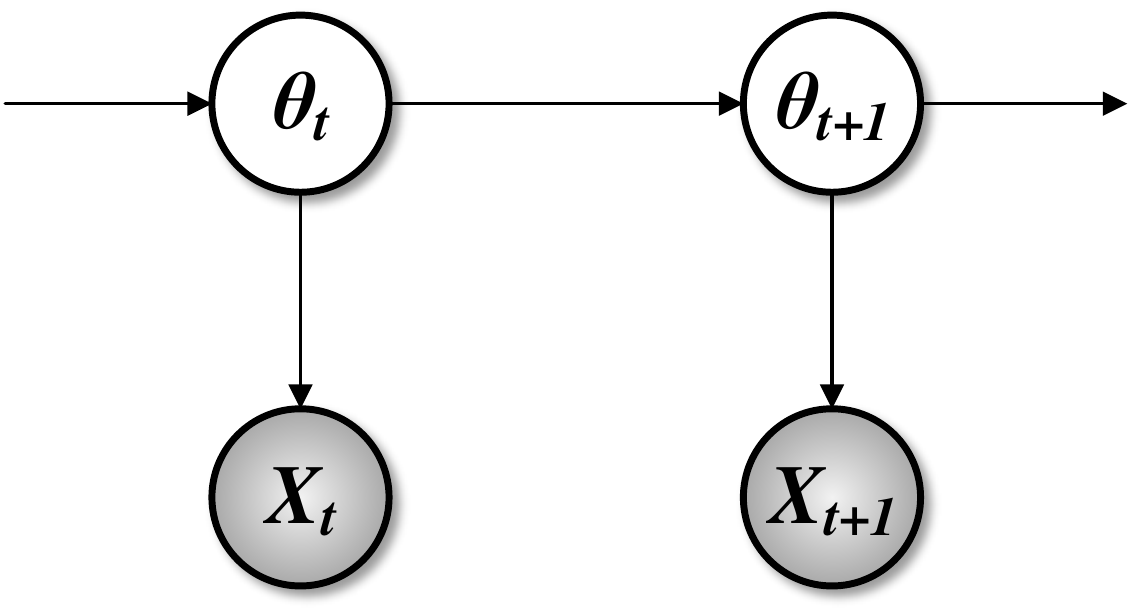}}
	\subfigure[Our approach for modeling group behavior.]{
		\label{newtra}
		\includegraphics[width=0.4\textwidth,height=3cm]{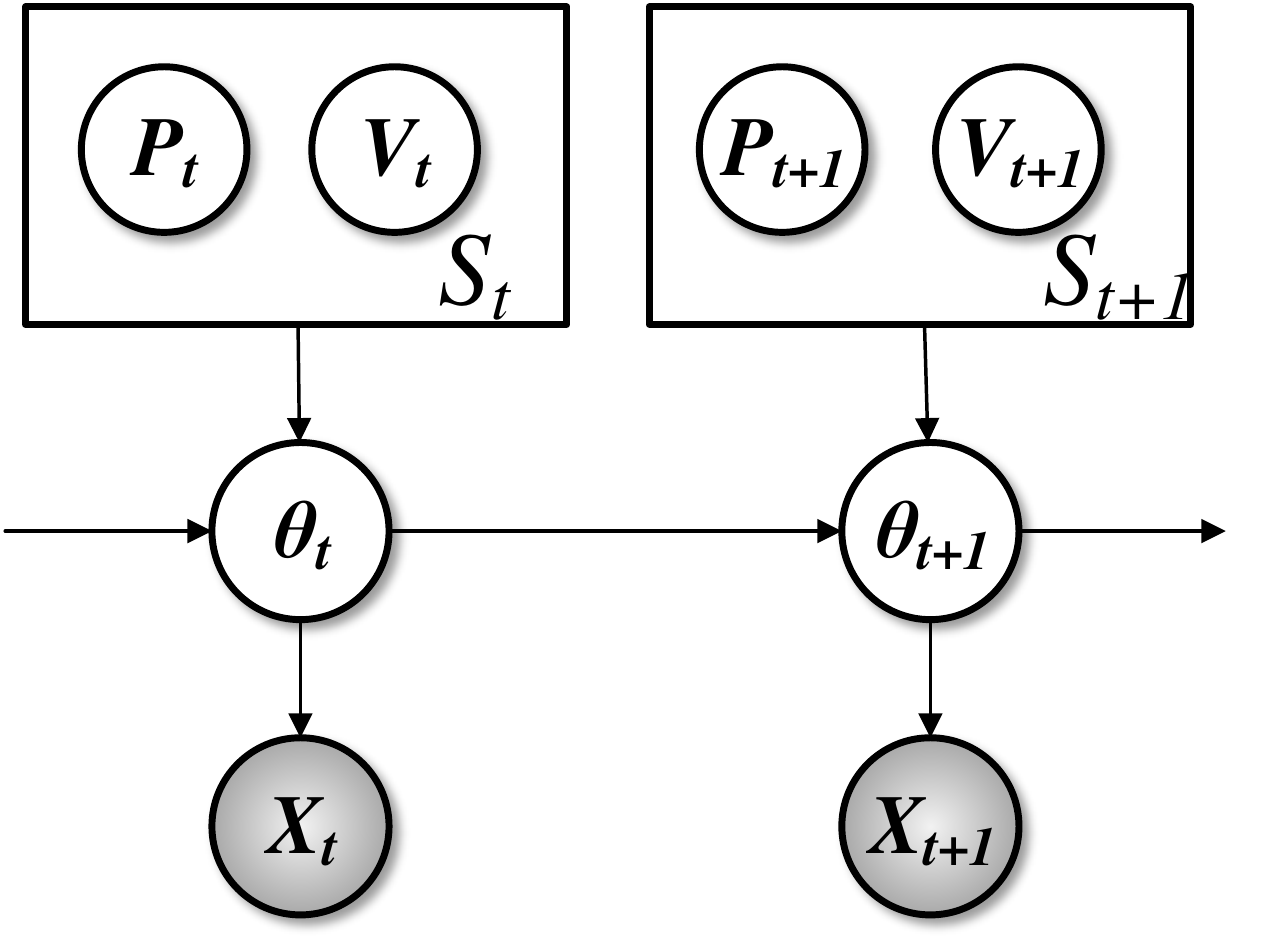}}
	\caption{Group Behavior Model.}
	\label{oldandnewtrack}
\end{figure*}
We first explain the difference between the proposed tracking model and existing ones. Fig. \ref{oldtra} represents the traditional state-space model employed for the generic object tracking. Formally, this state-space model is defined as follows:
\begin{equation} \label{statetransform}
	\begin{split}
		\theta_{t+1} = f_t(\theta_t), \\
		X_t = g_t(\theta_t),
	\end{split}
\end{equation}
where $\theta_t$ is the state of the target at time $t$, and $X_t$ is the observation. Thus, current state is only determined by the previous state, and $f_t$ and $g_t$ are nonlinear unknown functions.  Fig. \ref{oldtra} graphically describes the link from $\theta_t$ to $\theta_{t+1}$ and the link from $\theta_t$ to $X_t$  in Eq. \ref{statetransform}. Along this line of consideration, we model group behavior that includes surrounding information for tracking. In our group behavior model, the movement of each target is influenced by the neighboring ones.  Fig. \ref{newtra} illustrates our group behavior model and in this case Eq. \ref{statetransform} can be rewritten as follows:
\begin{equation} \label{ourtransform}
	\begin{split}
		\theta_{t+1} = f_t(\theta_t, S_{t+1}), \\
		S_t = (P_t, V_t),	\\
		X_t = g_t(\theta_t),
	\end{split}
\end{equation}
where $S_{t}$ is the surrounding information, including  the location $P_t$  and velocity $V_t$ information.

As we all know, vehicles have to  keep the minimum safe distance between each other   according to the traffic regulations. In other words, when two vehicles become too close, the back one will have the tendency that it will keep away from the front one. We assume that a potential repulsive force exists among vehicles in this situation and this potential repulsive force is named as traffic force (TF). The traffic force  makes each individual in the traffic scene  hold a minimum distance from others and avoid collision. We regard this behavior caused by the traffic force as GBM. By building the GBM, we try to simulate the   behaviors among vehicles and improve our tracker.

The traffic force between individuals is inversely related to their distance. If the distance decreases, this force increases. With this in mind, the distance between the predicted locations of targets can be used to calculate TF. Let $Tf_t = [tf_t^1, tf_t^2,...,tf_t^i,...]$ be the vector of traffic force. We use $tf^i_t$ to represent the TF  of the $i_{th}$  target from its neighbors . And the surrounding information $s_t^{ij} \in S_t$ is the $j_{th}$ target around the  target $i$. The overall force is defined as:
\begin{equation} \label{GBMOverall}
	tf_t^i=\sum_{i\not=j}{\mu_{ij}}w(s_t^{ij}),
\end{equation}
where $w(s_t^{ij})$ is the force between the target $i$ and its neighboring target $j$. Each TF between the two vehicles is computed as:
\begin{equation}
	w(s_t^{ij})=\exp(\frac{-d_{ij}^2(t)}{2\sigma_d^2}),
\end{equation}
where $\sigma_d$ controls the distances of a vehicle to be avoided, and $d_{ij}$ is the Euclidean distance between the two targets. We assume that target $i$ has the predicted position $prep_t^i$ and its neighboring target $j$ has the predicted position $prep_t^j$:
\begin{equation}
	d_{ij}(t)=\left \| prep_t^i-prep_t^j \right \|,
\end{equation}
Moreover, $\mu_{ij}$ in Eq. (\ref{GBMOverall})  represents the influence of  target $j$ in the overall TF on target $i$. It guarantees that  different vehicles will have different influences. The definition of $\mu_{ij}$ is:
\begin{equation}
	\label{influradius}
	\mu_{ij}=\exp(\dfrac{{-\left \| prep_{t-1}^i-prep_{t-1}^j \right \|}^2}{2\sigma_w^2}),
\end{equation}
where $\sigma_w$ is the radius of targets influence.

By utilizing the traffic force  $Tf_t$, we can take group behavior cue into consideration for tracking multiple vehicles. However, GBM here is different from social force model (SFM)\cite{09sfm} in theory. There are three main differences between GBM and SFM:
\begin{itemize}
	\item GBM is calculated based on the whole object, while SFM is calculated based on  pixels.
	\item GBM is defined by the distance, while SFM is based on the velocity.
	\item  GBM cares more about the movement of the individual target affected by the whole group. SFM, on the other hand, focuses on the movement tendency of the whole group.
\end{itemize}

\subsubsection{GBM-Based Tracking}
Our tracking model utilizes  GBM  in the Kalman filter \cite{Krebs76b} to predict the locations of vehicles. For the state prediction, our constraint of group behavior model is applied in the following form:
\begin{equation} \label{eqkalman}
	\theta_{t}=Tf_t \cdot F_t \cdot \theta_{t-1} + B_t \cdot u_t,
\end{equation}

where $F_t$ is the process transition matrix, $u_t$ is the control vector, and $B_t$ converts control vector $u_t$ to state space.  The state $\theta_{t}$ will be predicted by $\theta_{t-1}$ under the constraint of $Tf_t$. Recalling the definition of $w(s_t^{ij})$ in \ref{GBMSection}, it is easy to know that $0<w<1$. The physical significance of $w$ can be seen as decelerating vehicles that are affected in the group. And this  constraint is reasonable due to the traffic regulations, in which   to keep the minimum safe distance between vehicles and avoid collisions, vehicles have to slow down.  From this perspective,  the modification to the Kalman filter is promising. After obtaining predicted locations, these locations and new detected ones will be assigned to existing trajectories. We assign locations to trajectories by judging the motion tendency of a vehicle. If one location is consistent with the motion tendency and it is  near the trajectory, the location will be allocated to the trajectory and the trajectory will be updated.  A simple greedy algorithm will be employed to assign locations to  trajectories.   If one location doesn't belong to any trajectory. It will be the start of a new trajectory.

The reasons why the Kalman is chosen are as follows: First, Kalman filter has less computation price than other methods. Second, vehicle velocity  will not change violently in this traffic context. Because of traffic light, most   vehicles have similar velocities. Thus, velocity has a Gaussian distribution that is the necessary condition for Kalman filter. Therefore, Kalman filter is suitable for this situation.

The whole tracking-by-detection procedure is summarized in Algorithm \ref{algorithm1}.
\begin{algorithm}
	\caption{GBM-based Tracking algorithm} \label{algorithm1}
	\begin{algorithmic}
		\Procedure {GBMTracker}{$videoseq$}
		\State 1.~Initialize tracker and detector
		\State 2.~Get each $frame$ from $videoseq$
		\State 3.~Obtain possible locations as many as possible via DPM with $\eta=-0.78$
		\State 4.~Extract vehicles from possible locations via shape segmentation Eq. \ref{energy}-Eq. \ref{endseg}
		\State 5.~Modeling group behavior via detection information Eq. \ref{ourtransform}-Eq. \ref{influradius}
		\State 6.~Predict locations Eq. \ref{eqkalman}
		\State 7.~\textbf{if} $new ~location ~belongs ~to ~existing ~trajectory$
		\State ~~~~ update assigned tracks
		\State ~~~\textbf{else}
		\State ~~~~ generate new tracks
		\State ~~~\textbf{end if}
		\State 8.~Display result
		\EndProcedure
	\end{algorithmic}
\end{algorithm}

\section{Experiment and Discussion} \label{experiment}

%\begin{figure*}
%  \centering
% \includegraphics[width=1\textwidth]{a}
%\caption{\label{result} Exhibit the detection result compared with original DPM. The frame %comes from real surveillance video of the traffic junction. The left is the result of original %DPM with normal threshold value. The middle one shows the result of original DPM with %significant threshold value. The right figure demonstrate our outstanding performance of %hierarchical detection model. }
%\end{figure*}

To demonstrate the capabilities of the presented approach,   extensive experiments are conducted and evaluated. In this section,  the experiments will be introduced from the following aspects: data set, evaluation measure, parameter selection,   experimental results and analysis.

\subsection{Data Set}
Multiple object tracking has many public data sets. However, there are no surveillance videos in the road junctions. For this reason, our experiments are performed on videos that we collected. The  data set contains one short video (Seq1) and two long videos (Seq2 and Seq3).   Seq1 involves 748 frames (688$\times$384) and  25 frames per second. It describes only two opposite directions.   Seq2 includes 6200 frames (1280$\times$720), while   Seq3 includes 7908 frames (1280$\times$720).   Both of them are 30 frames per second and describes all the possible directions in road junctions.

\subsection{Evaluation Measure}
Diverse evaluation measures are employed for different stages in our approach.
\subsubsection{Evaluating Detection}
Precision-recall measure is adopted to evaluate the detection performance. Precision (also called positive predictive value) is the fraction of retrieved instances that are relevant, while recall (also known as sensitivity) is the fraction of relevant instances that are retrieved. For classification tasks, the terms true positives (TP), true negatives (TN), false positives (FP) and false negatives (FN) compare the classifier results   under test with trusted external judgements. Based on these definitions, the two metrics are calculated as:
\begin{equation}
	precision = \frac{TP}{TP + FP},
\end{equation}
\begin{equation} \label{recalleq}
	recall = \frac{TP}{TP + FN},
\end{equation}
In our vehicle detection task, TP is the number of vehicles that are correctly detected on all frames. FP is the   ones that are incorrectly detected as positives and  FN is the ones that are not detected but should have been detected.

\subsubsection{Evaluating Tracking}
We evaluate our tracking results using the standard CLEAR MOT metrics \cite{clearmot}. The indexes of CLEAR MOT metrics are MOTP (multiple object tracking precision) and MOTA (multiple object tracking accuracy). MOTP shows the ability of the tracker to estimate precise object positions, independent of its skill at recognizing object configurations, keeping consistent trajectories, and so forth. Briefly, MOTP embodies the precision of the target locations. It is defined as:
\begin{equation}
	MOTP = \frac{\sum_{i,t}d^i_t}{\sum_tc_t},
\end{equation}
where $\sum_{i,t}d^i_t$ is the total error in estimated position for matched object-hypothesis pairs over all frames, and $\sum_tc_t$ is the total number of matches made.

Moreover, MOTA accounts for all object configuration errors made by the tracker, false positives, misses, mismatches, over all frames. Compared with MOTP, MOTA cares more about the accuracy of the number of targets.  MOTA can be seen as derived from three error ratios:
\begin{equation}
	MOTA = 1 - \frac{\sum_t(m_t+fp_t+mme_t)}{\sum_tg_t},
\end{equation}
where $\sum_tm_t / \sum_tg_t$ is the ratio of misses in the sequence, $\sum_tfp_t / \sum_tg_t$ is the ratio of false positives and $\sum_tmme_t / \sum_tg_t$ is the ratio of mismatches. We count all tracker hypotheses for which no real object exists as false positives and count all occurrences where the tracking hypothesis for an object changed compared to previous frames as mismatch errors.   All of them are computed over the total number of objects present in all frames.

\subsection{Parameter Selection}
\label{choosepara}
In our framework, some parameters play an important role in the experiments. In order to obtain a significant performance, the value of those parameters should be selected meticulously.
\begin{figure}
	\center
	\includegraphics[width=0.35\textwidth,height=3cm]{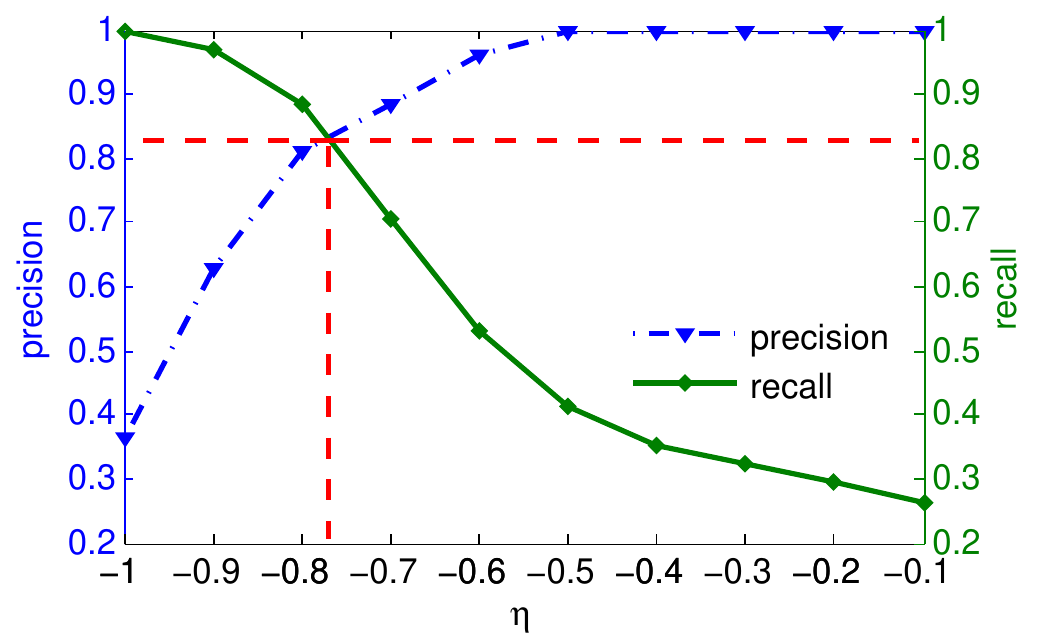}
	\caption{\label{para1} Effect of varied $\eta$ on the precision and recall results.}
\end{figure}
\begin{figure}
	\center
	\includegraphics[width=0.35\textwidth,height=3cm]{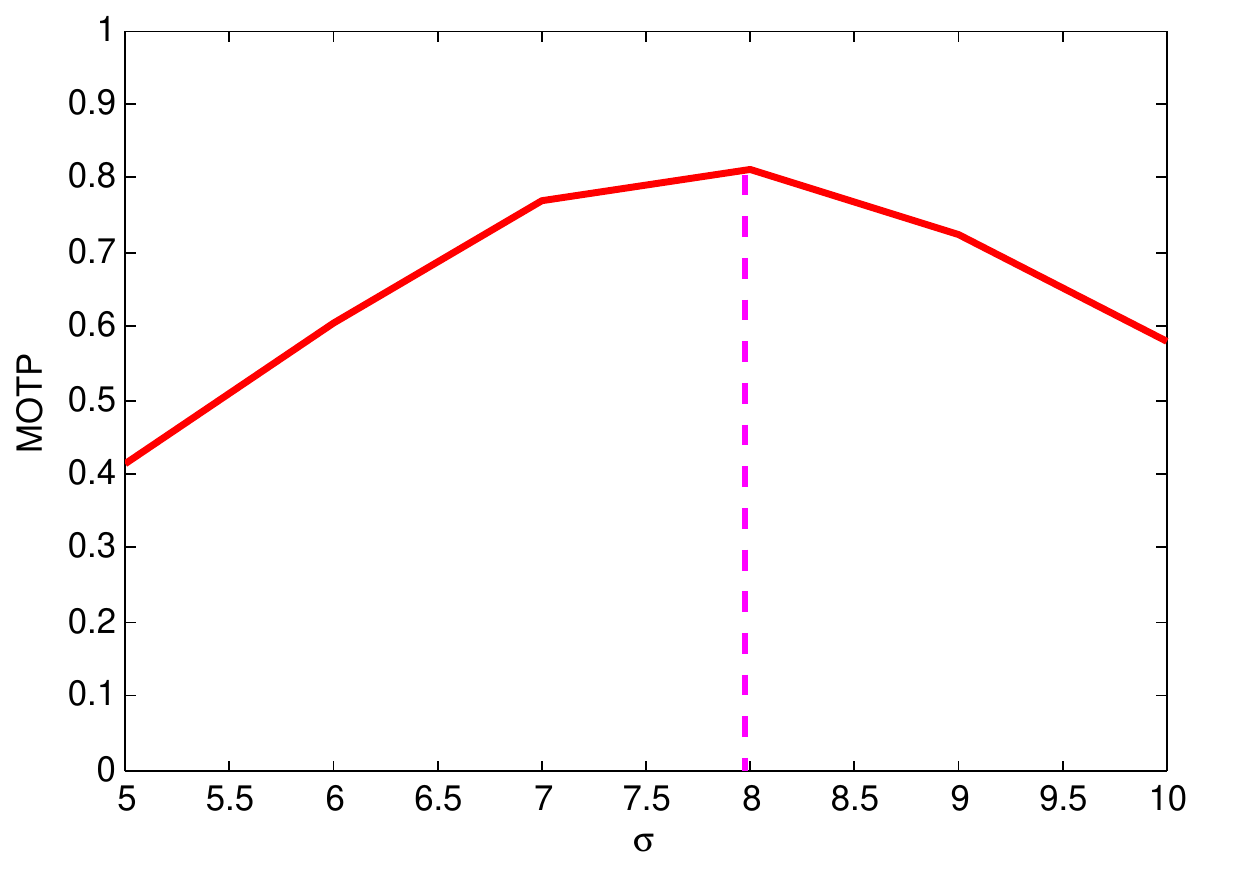}
	\caption{\label{para2} Effect of varied $\sigma_w$ on the MOTP results.}
\end{figure}
\subsubsection{Threshold $\eta$ in Detection}
As mentioned before, DPM in our framework is used to obtain vehicle hypotheses as many as possible. To this end, we are supposed to choose a much smaller threshold $\eta$ that is used to determine whether the pixel area is object or not. However, the default $\eta$ equals to -0.5. This threshold cannot provide enough possible hypothesis in road junctions. Therefore, we select another proper threshold through experiment.

\begin{figure*}
	\centering
	\subfigure[DPM detector]{
		\label{olddet}
		\includegraphics[width=0.35\textwidth,height=3cm]{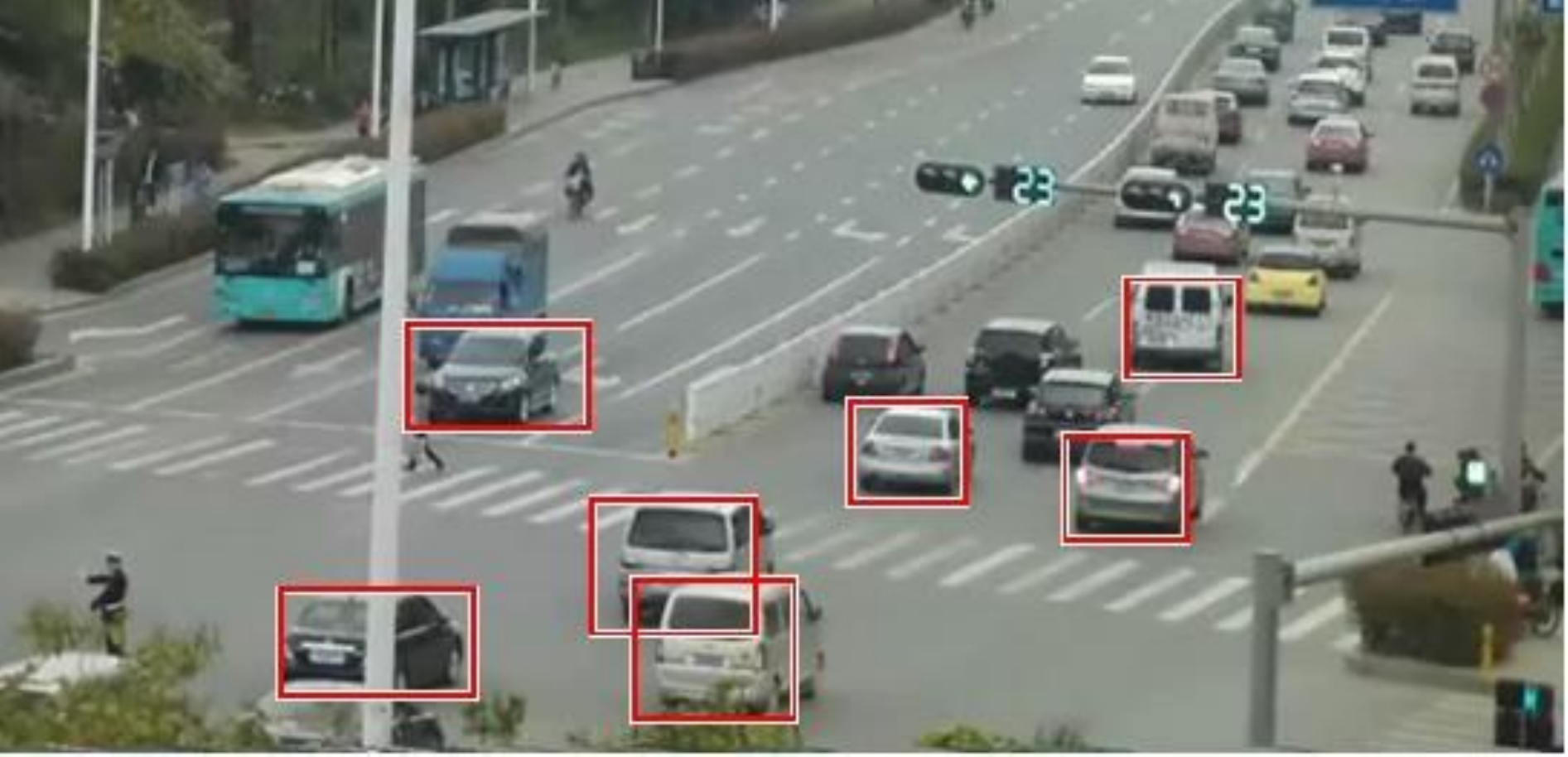}}
	\subfigure[Detecting with shape prior]{
		\label{ourdet}
		\includegraphics[width=0.35\textwidth,height=3cm]{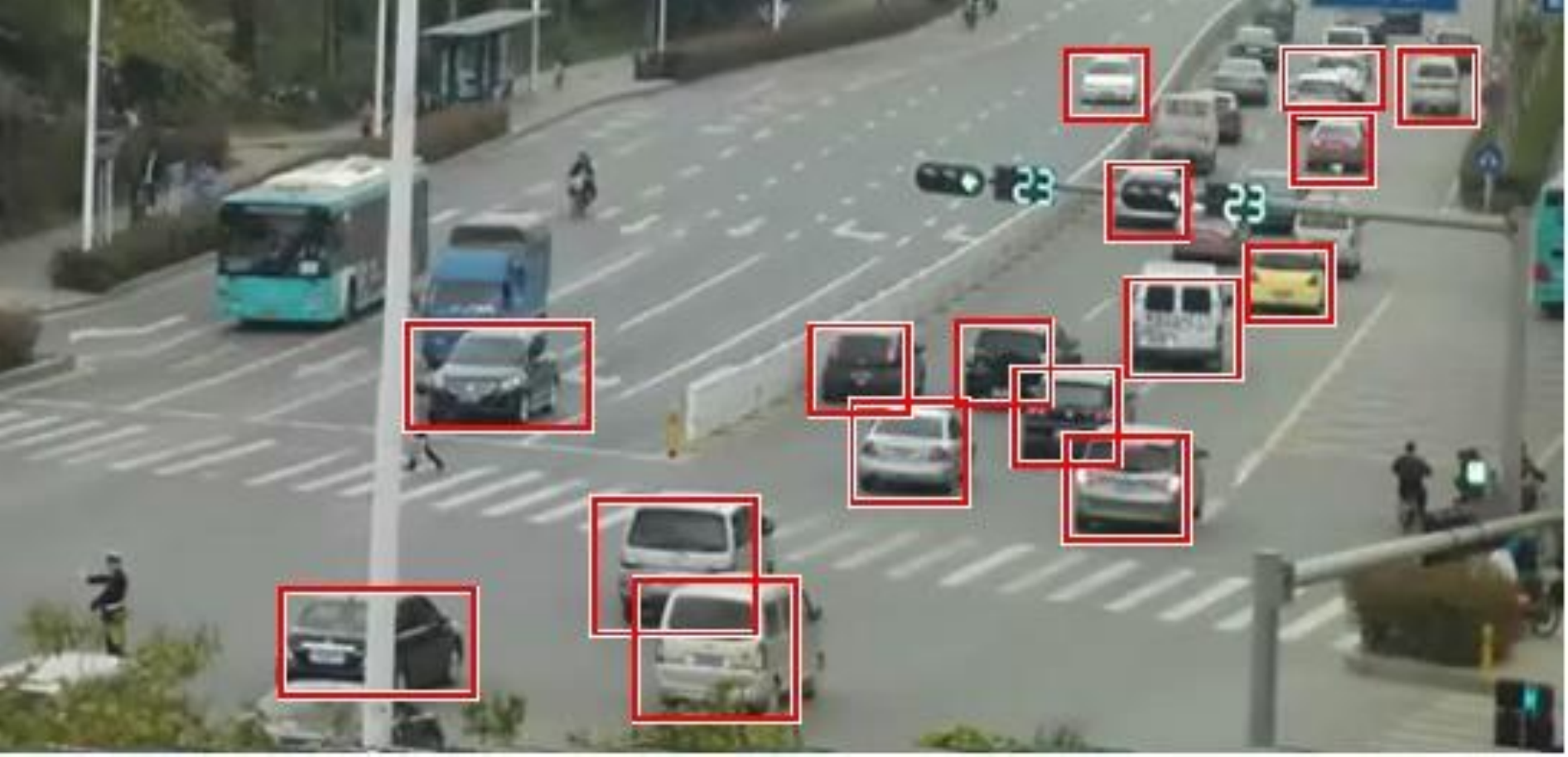}}
	\subfigure[DPM detector]{
		\label{chongdie}
		\includegraphics[width=0.35\textwidth,height=3cm]{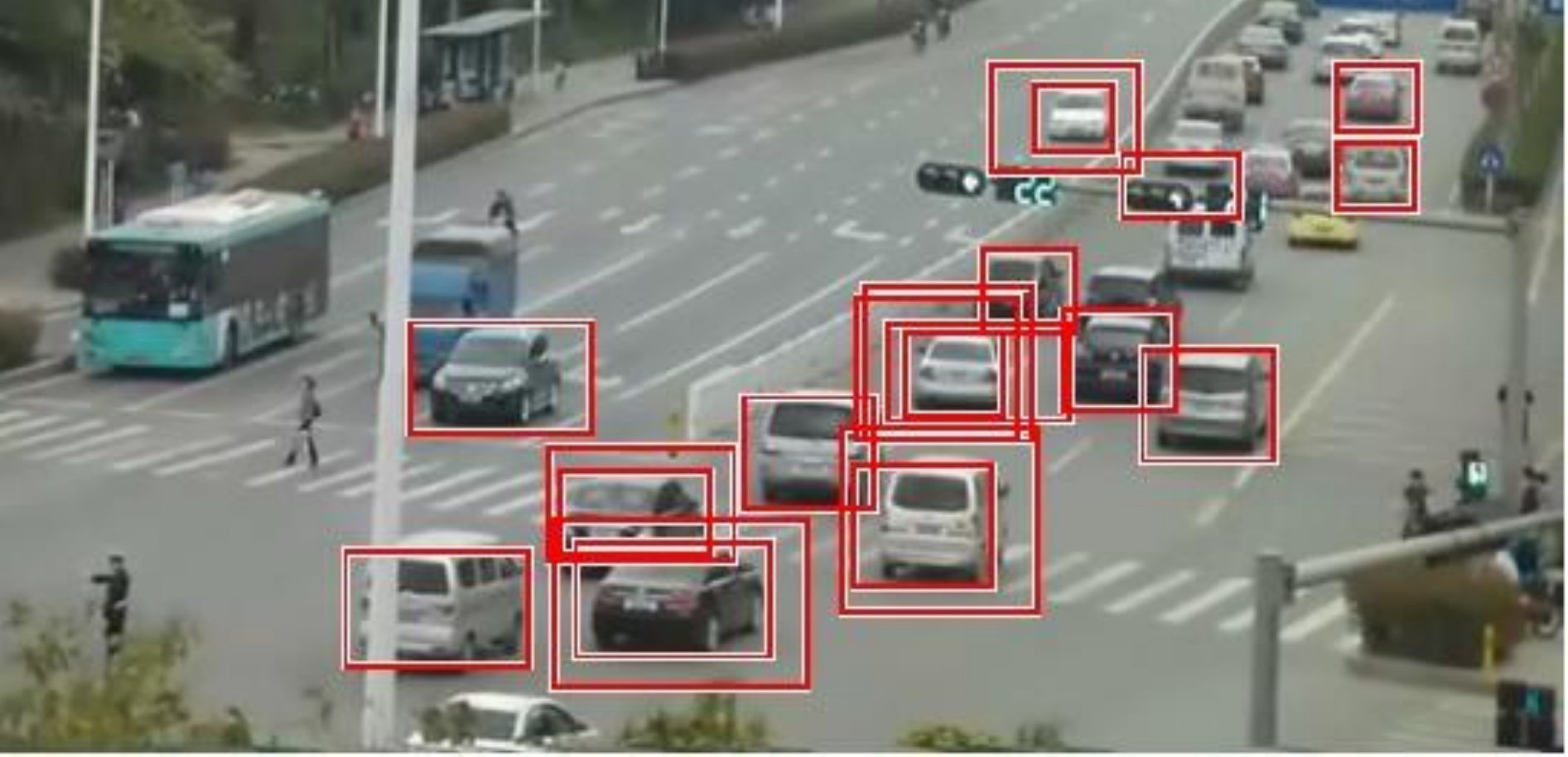}}
	\subfigure[Detecting with shape prior]{
		\label{quchong}
		\includegraphics[width=0.35\textwidth,height=3cm]{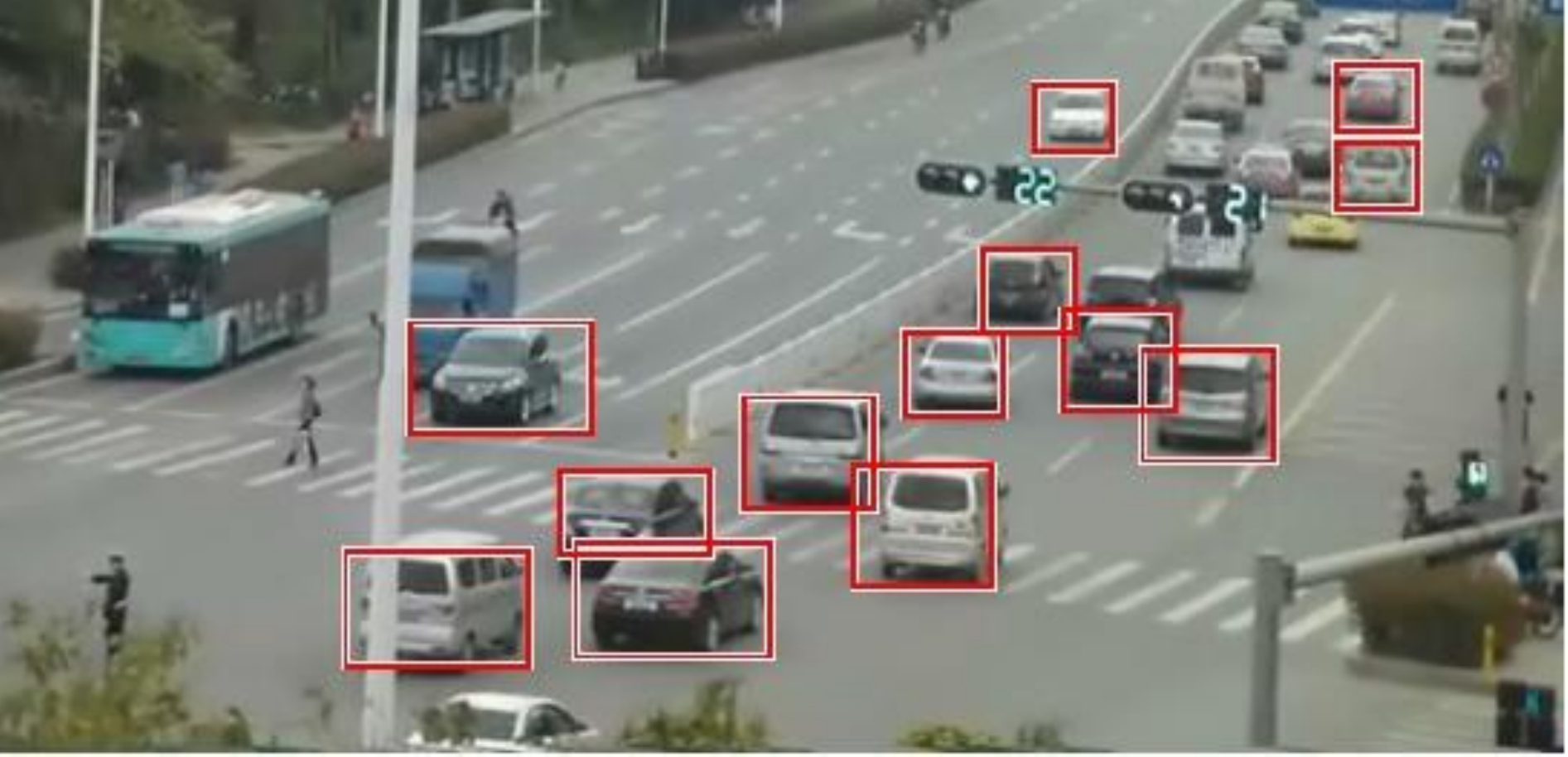}}
	\caption{The first column is the detection result with original DPM detector, while the second column shows the result of the proposed sequential detector with shape prior.}
	\label{detshow}
\end{figure*}

Fig. \ref{para1} displays how  $\eta$ is selected. As we all know that the  precision is inversely related to the recall. It is impossible to make precision and recall the biggest simultaneously. In Fig. \ref{para1},  we can see that when $\eta$ is smaller than $-0.5$,  recall of our detector increases significantly, but precision decreases slowly. With the recall improving, false detections should have increased and the precision should have decreased greatly. However, the reason why the precision decreases slowly is that our shape prior segmentation that is added into the sequential detection model prevent it from being in steep decline. According to the shape difference between vehicles and other target, shape prior segmentation takes out many false detections which is produced by a significant low threshold value $\eta$. Proper exploring the shape prior segmentation  makes us seek out a point that has relatively high precision and recall at the same time. Therefore, we finally choose $\eta=-0.78$.

\subsubsection{Influence Radius $\sigma_w$ of Each Target}
\label{sigmasec}
In Section \ref{socialcue}, each vehicle in our tracking model has its own influence radius. In theory, each vehicle will be affected by its neighboring ones  from all directions. However, due to the limitation of viewpoint, the influence from various direction seems different.  Given this actual situation, our influence radius of vehicles is selected by experiment.

Fig. \ref{para2} shows our experimental curve of selecting $\sigma_w$. With the increase of $\sigma_w$, the MOTP of our tracker improves first but decreases after the peak value. We attempt to explain the  result through the phenomenon of the realistic traffic scene: When the influence area of a vehicle is too small, the relationship between vehicles become weak. Our tracker will degenerate into a conventional tracker that only consider the vehicle itself. This will lead to a poor performance.    On the contrary, if a vehicle influences much large areas, more vehicles will be affected. That is to say,  not only the neighboring ones are affected, the others are also  influenced. This will be contradictory to our assumption that vehicles can only affect their neighboring ones. The performance is also unsatisfying.  Hence, we finally select $\sigma_w=8.0$ for our group behavior model.

\subsection{Result exhibition}
With the parameters presented at section \ref{choosepara}, a more detailed analysis of our experiment will be presented in the following sections respectively.
\subsubsection{Sequential Detection with Shape Prior}

Employing shape prior segmentation in detection makes our detector unique. Fig. \ref{detshow} reveals our experiment results of sequential detection. Fig. \ref{olddet} employs original DPM with its common threshold $\eta=-0.5$, while Fig. \ref{ourdet}, Fig. \ref{chongdie} and Fig. \ref{quchong} utilize our detector that has a lower threshold $\eta=-0.78$ with shape information of vehicles. When not applying shape information of vehicles, we will get fewer targets in Fig. \ref{olddet} or more overlapping bounding boxes in Fig. \ref{chongdie}. After making use of shape prior, Fig. \ref{ourdet} demonstrates we can obtain more targets, while Fig. \ref{quchong} shows another superiority of shape prior that can remove false detections. 

The advantages of our detecting method are as follows: First, we use a significant low threshold value. In original DPM, this low threshold value will lead to lots of false detections and duplicate bounding boxes. However, we don't have this trouble. We just regard the output of DPM as proposals, while they are final detection results in original DPM. With this low threshold value, we can get candidates of vehicles as many as possible without considering the false detection. Second, shape prior is added to the segmentation. The significant low threshold value allows us to detect more targets in theory, while the shape prior segmentation guarantees the feasibility of detecting more targets in practice. We employ this shape prior segmentation to deal with the proposals obtained by DPM. Owing to applying shape information of the vehicle, we can easily distinguish correct vehicles from false detections.

\begin{figure*}
	\centering
	\subfigure[Comparison with ROC curve.]{
		\label{roc}
		\includegraphics[width=0.35\textwidth, height=3cm]{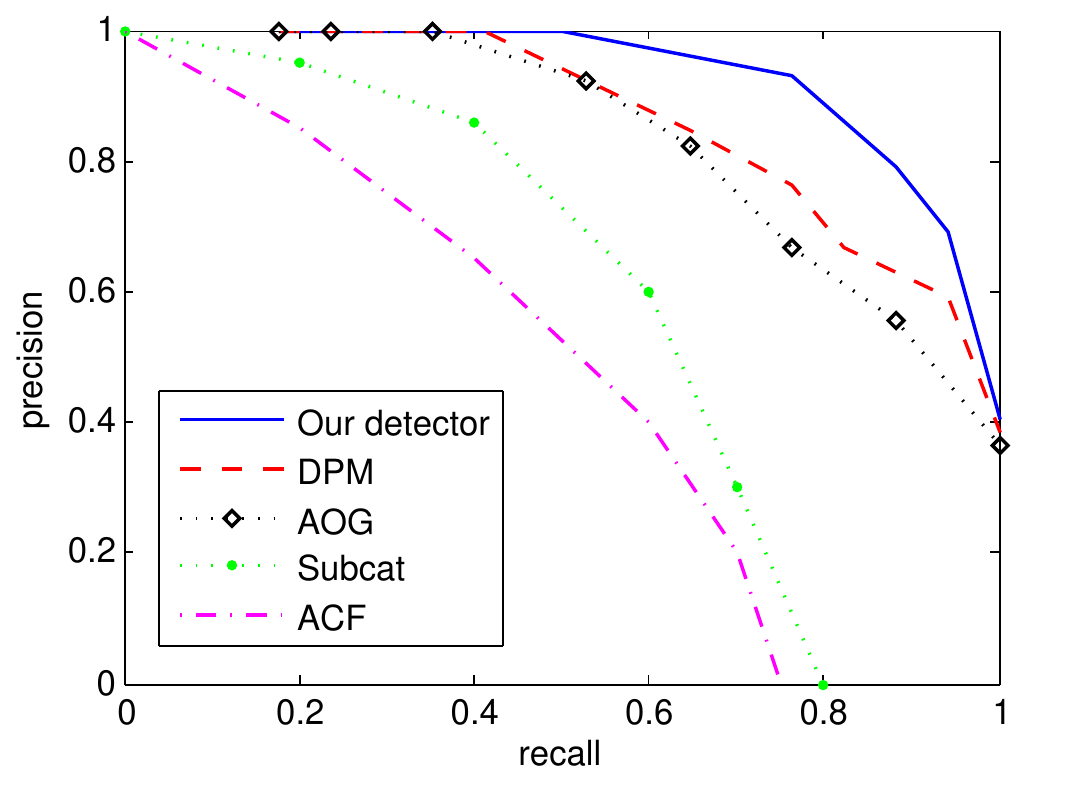}}
	\subfigure[Comparison with FPPI curve.]{
		\label{fppi}
		\includegraphics[width=0.35\textwidth, height=3cm]{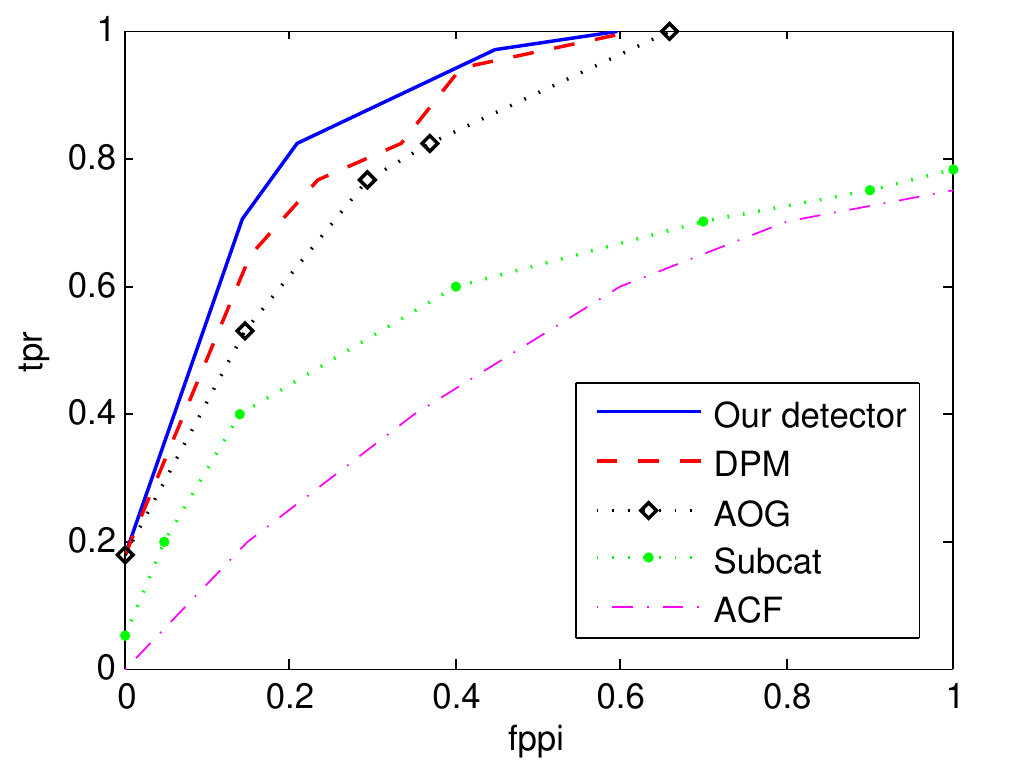}}
	\caption{Comparison between five detection methods, including our sequential detector, DPM detector, AOG detector, original ACF detector and Subcat detector.} 
%		which is a ACF detector with classifying vehicles into subcategories. }
\end{figure*}

We compare our detector with DPM \cite{DPMpami}, AOG \cite{eccvLiWZ14} and ACF-based  detectors (Subcat \cite{ohnbar14} and ACF \cite{DollarPAMI14pyramids})  . Both DPM and AOG deal with occlusion very well, while ACF (Aggregate Channel Features) is popular in recent years because of its speediness and high efficiency. Subcat make  vehicle detection much faster, due to classifying vehicles into subcategories.  Fig. \ref{roc} and Fig. \ref{fppi} shows the result of comparison. ROC (receiver operating characteristic) is a graphical plot that illustrates the performance of a binary classifier system as its discrimination threshold is varied. FPPI represents false positives per image and TPR (true positive rate) has the same value with recall in Eq. \ref{recalleq}.

%eccvLiWZ14

Nevertheless, different from results in public data set, ACF-based detectors \cite{ohnbar14,DollarPAMI14pyramids} have a poor performance in road intersections. The reason is that ACF detectors doesn't specially aim at occlusion that is quite serious in our traffic surveillance video. On the other hand, AOG and DPM has similar performance taking occlusions into account. And our detector, which divide detection problem into detecting and segmentation, own the best performance in such a real scene, since we have more candidates and make full use of shape information of the vehicle. The result also proves our thought that aggregating several basic techniques can reach a significant performance.

\begin{figure*}
	\centering
	\subfigure[ frame\#161]{
		\label{drifta}
		\includegraphics[width=0.35\textwidth,height=3cm]{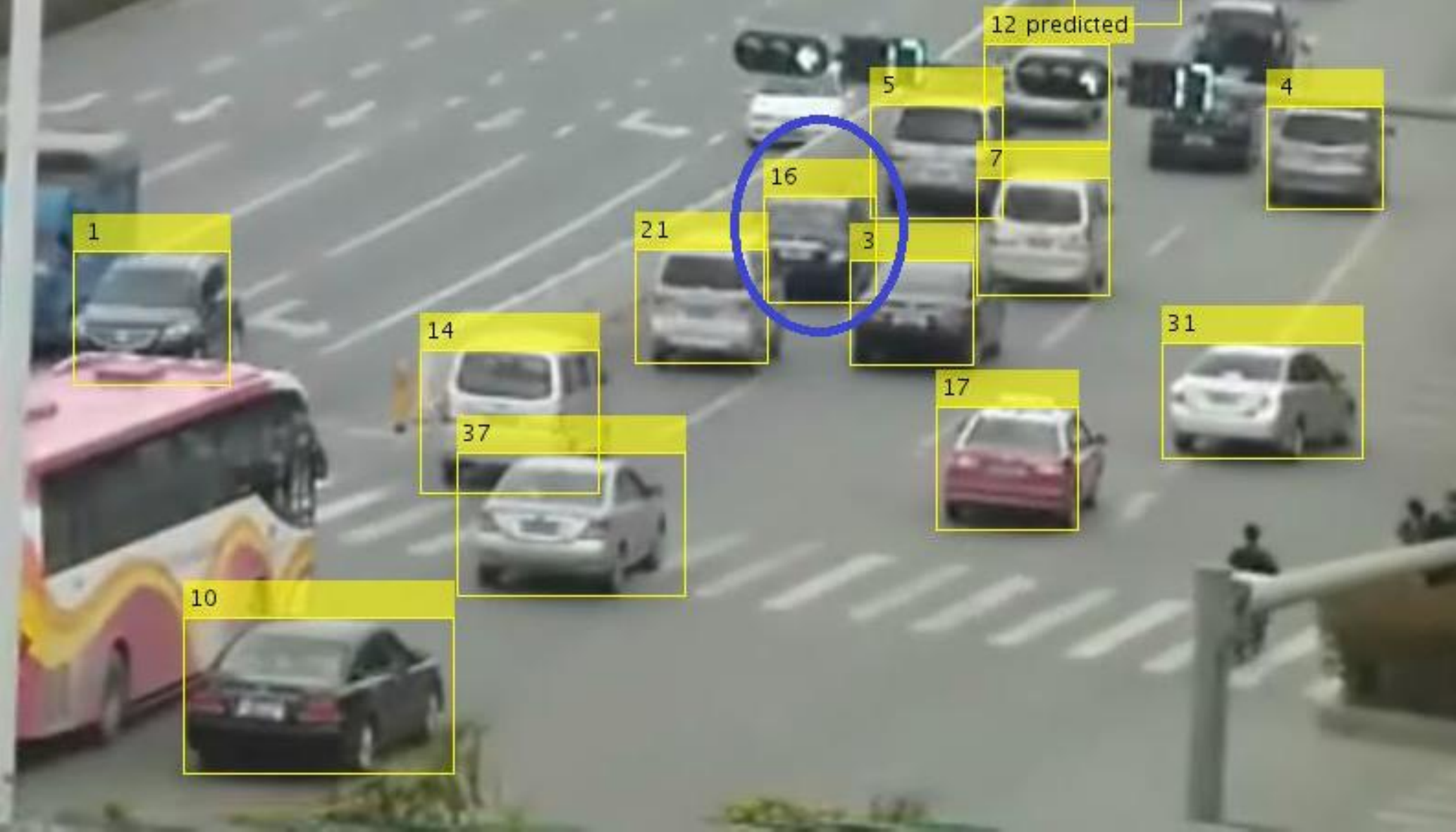}}
	\subfigure[ frame\#171]{
		\label{driftb}
		\includegraphics[width=0.35\textwidth,height=3cm]{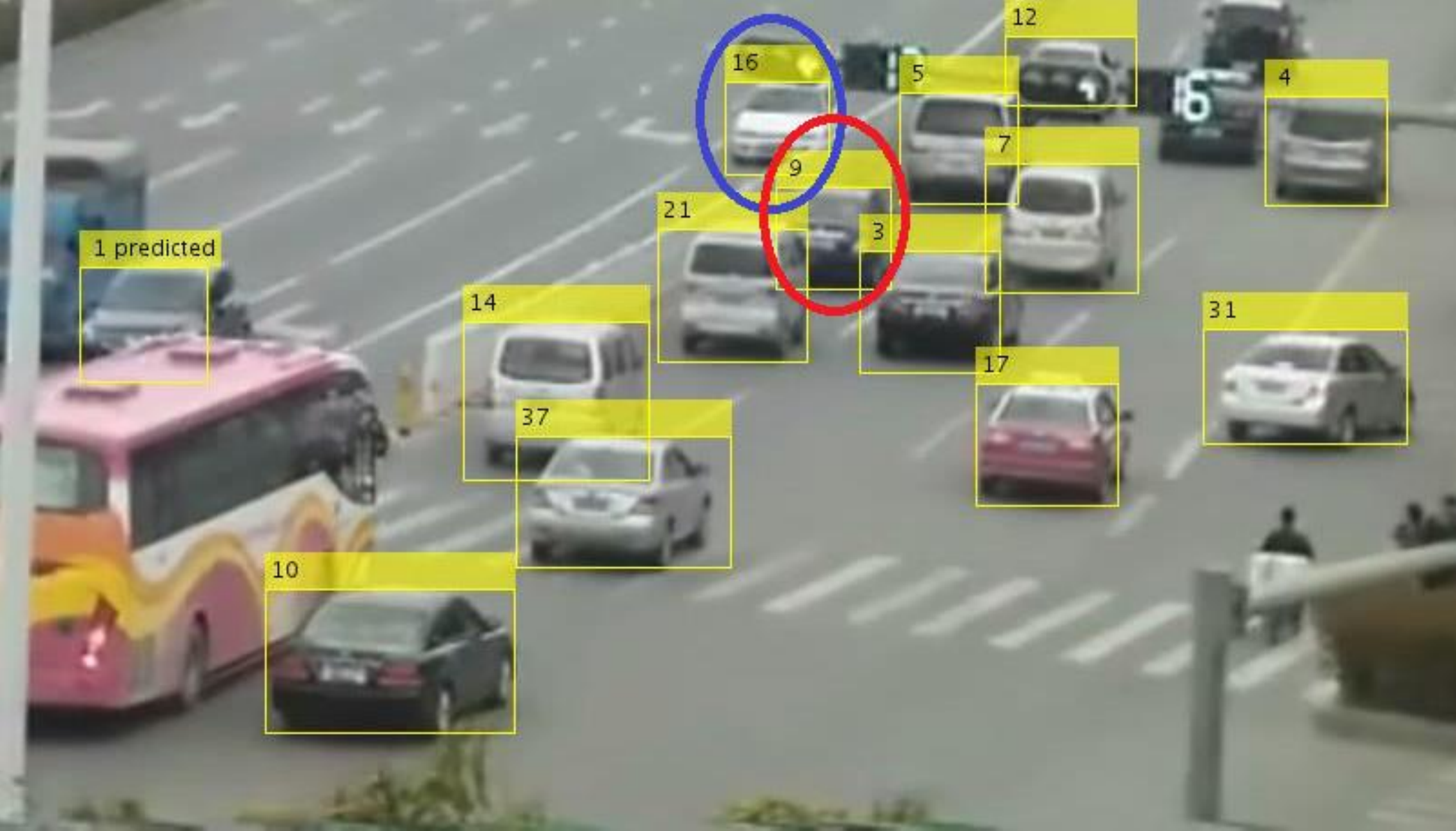}}
	\subfigure[ frame\#161]{
		\label{driftc}
		\includegraphics[width=0.35\textwidth,height=3cm]{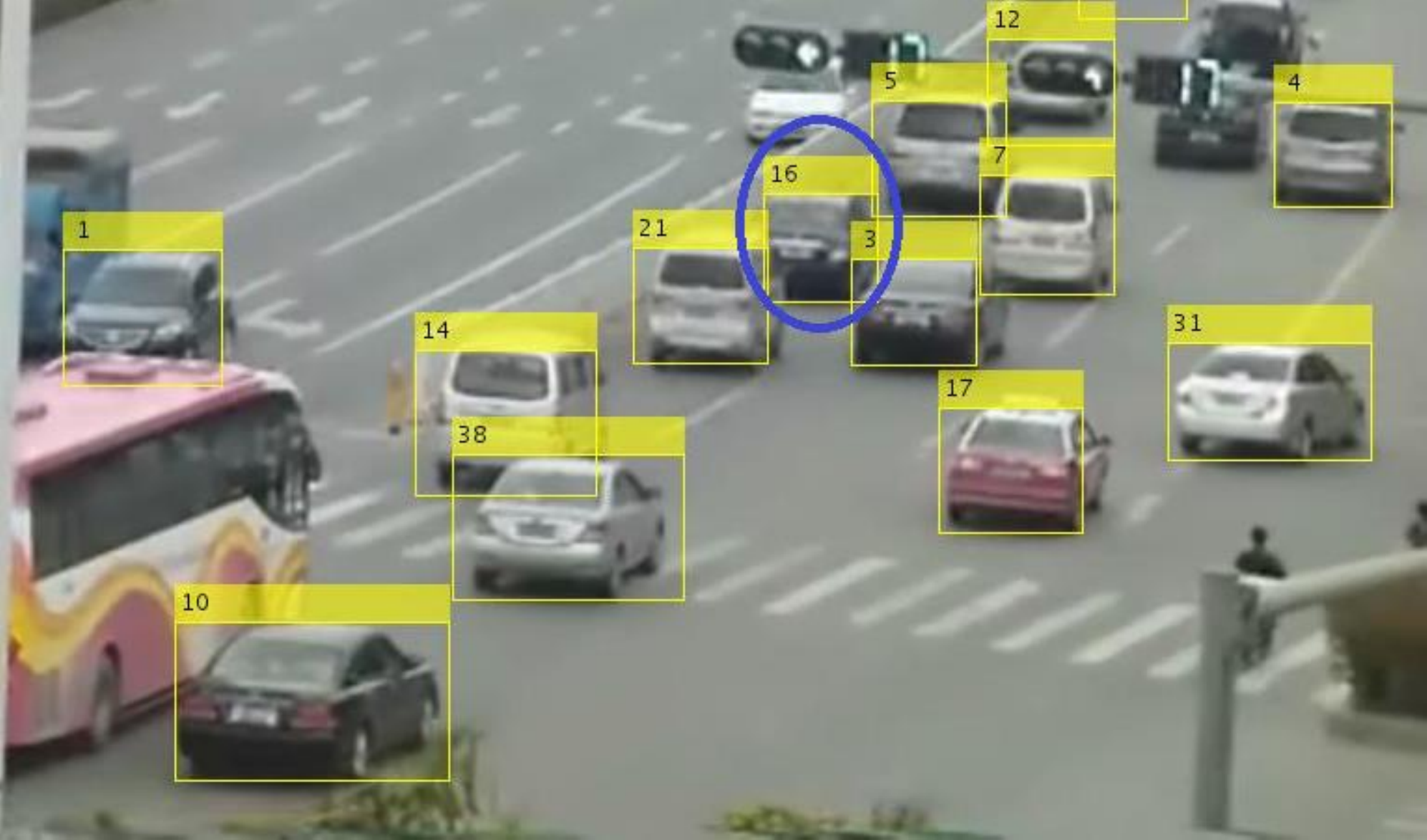}}
	\subfigure[ frame\#171]{
		\label{driftd}
		\includegraphics[width=0.35\textwidth,height=3cm]{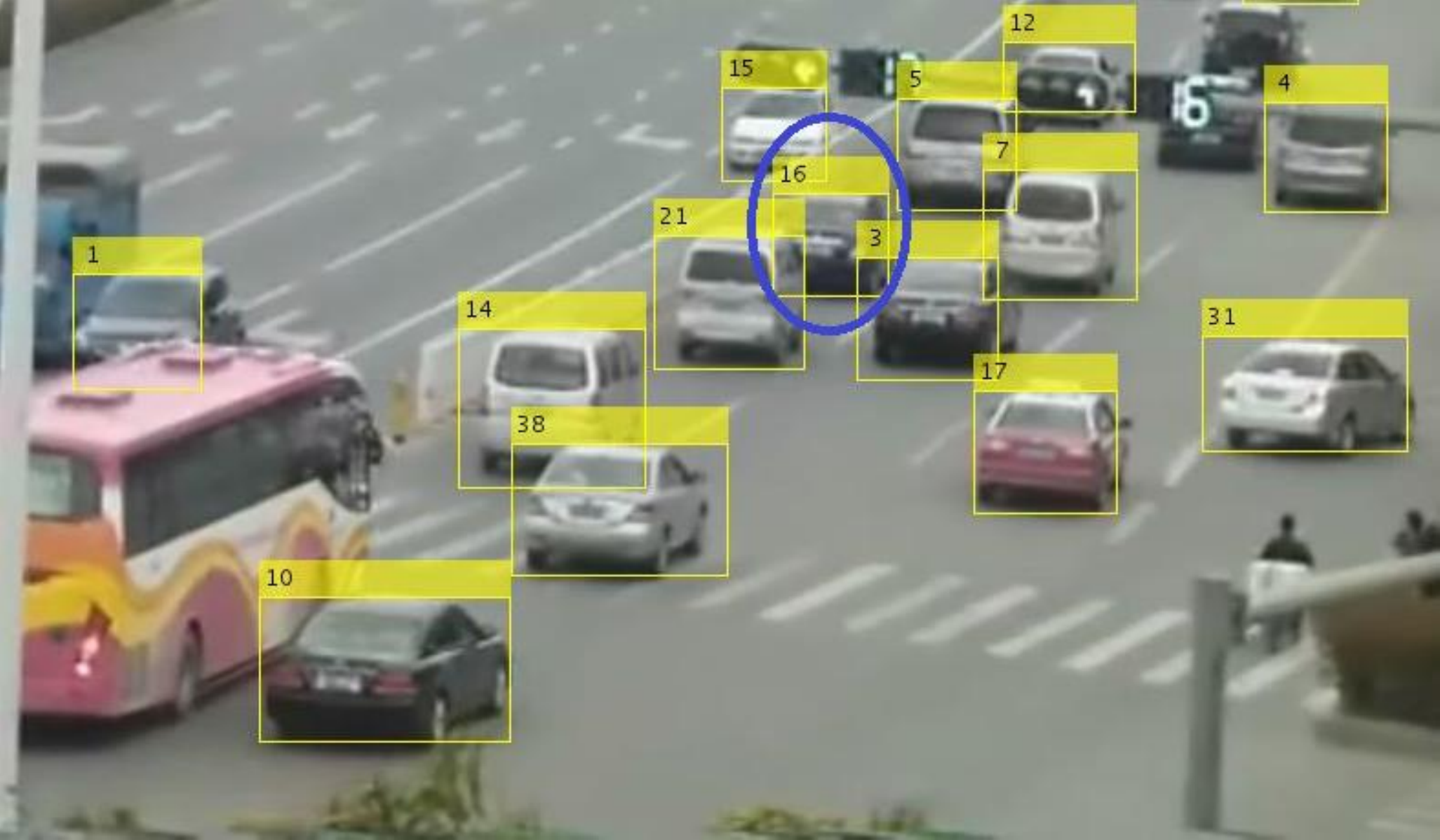}}
	\caption{The first row is the tracking result without modeling social behavior, while the second row shows the result   with GBM. Different number of the bounding box indicates the different vehicles.}
	\label{drift}
\end{figure*}

Despite that the performance of our sequential detection is outstanding. There are still some problems to be discussed. One of the controversial problems is that whether we should detect all the vehicles in a frame or not. That is to say, for those distant targets whose size is quite small, is it necessary to detect them all? Early in our research, we tried to detect all the vehicles regardless of their size. Unfortunately, it is really difficult to detect all the vehicle, especially for tiny ones. They are even hard to identify the structure. Therefore, we only focus on those vehicles near the intersection. When those distant vehicles come to near, we can also detect them. On the other hand, the traffic junction is an accident black spot. Detecting those distant vehicles seems to be meaningless. Moreover, those distant vehicles may be near ones compared to other camera. In consequence, we only detect those near target in this work. It is sufficient for surveillance video of road junction. Another controversial problem is that since the templates of shape prior segmentation are from the training of DPM, will the results from DPM and shape prior segmentation be correlated and detection errors be reinforced? Though the DPM models  are correlated to the templates of shape prior segmentation, the detection errors cannot be reinforced. DPM models are more concerned about the relationship between the various parts of the object, however, shape templates only care about the similarity of the whole object. They influence detection results in different aspects. Fig. \ref{chongdie} and Fig. \ref{quchong} also prove this point of view. The traffic light detected in Fig. \ref{chongdie} is removed by shape prior segmentation in Fig. \ref{quchong}.

\subsubsection{GBM-Based Tracking}

Modeling group behavior of vehicles contributes to tracking in road junctions and taking interplay of targets into consideration makes our result more reasonable. Fig. \ref{drift} reveals one of the advantages of modeling group behavior, which can relieve the drift of targets. In Fig. \ref{drifta} and Fig. \ref{driftb}, the tracker regards  bounding boxes in blue circle as the same vehicle. Obviously, they are different targets and the vehicle's number in the red circle changes from 16 to 9. That means drift happens during the tracking course. After modeling group behavior in our tracking model, we no longer track a vehicle separately. Each vehicle's state is influenced by its surroundings which is relatively stable in the actual situation of real world. Therefore, in Fig.\ref{driftc} and Fig. \ref{driftd}, we can see that the vehicle in blue circle keep the same number. In our GBM, when interaction occurs between vehicles, GBM restricts  each target in the group to follow its trajectory. Since each vehicle is restrained by its surroundings, drifting becomes difficult.

By the same token, another advantage of GBM is that our tracker is better suited to complex movement of vehicles. Interplay between vehicles can be ignored in some simple scene. But for the traffic scene,  it is impossible. When a vehicle crosses the view of camera, the movement of its surroundings will also be taken into account. Thus, we treat these vehicles as a group. They can not only affect each other, but also have their own trajectory.

Since methods of multi-target tracking by minimizing energy function are popular in recent years, we compare our result with some typical methods CEM \cite{MilanRS14pami} and DCO \cite{SR12}. The baseline tracking results are from Kalman Filter(without the traffic force).  CEM focus on designing an energy that corresponds to a more complete representation of the problem, rather than one that is amenable to global optimization. It takes into account physical constraints, such as target dynamics, mutual exclusion, and track persistence.  In addition, partial image evidence is handled with explicit occlusion reasoning, and different targets are disambiguated with an appearance model in CEM. On the other hand, DCO proposes a discrete-continuous optimization method for minimizing energy function. In DCO, data association is performed using discrete optimization with label costs, yielding near optimality. And trajectory estimation is posed as a continuous fitting problem with a simple closed-form solution, which is used in turn to update the label costs. Due to not need to pre-compute trajectories, the accuracy of estimating trajectories improves.

Although CEM and DCO have great performance in some public data sets, they still handle targets individually. The relationship between targets is ignored. Compared with these  trackers, our GBM model group behavior to handle interactions among targets. GBM will help to predict locations more accurately in traffic video due to considering both vehicles and their surroundings. TABLE \ref{tab} shows the final results of the experiments, while TABLE \ref{tabmotp} and TABLE \ref{tabmota} exhibit the MOTP and MOTA in three test sequences, respectively.    In GBM, each target belongs to a group. It  will be affected by group members. The influence between vehicles prevents them from drifting, and it makes targets follow regular motion model. However, both CEM and DCO haven't applied group information. They just tracking vehicles individually. Therefore, our MOTP value is obviously higher. We have to admit that our MOTA is not superior. Since there are many  targets in traffic video sequence, improving accuracy without any other modifying of tracker is difficult. This problem maybe remit in our future work.

\begin{table}[!hbp]
	\center
	\begin{tabular}{c | c | c | c}
		\hline
		{\bf Tracking Algorithm} & {\bf Seq1}  & {\bf Seq2} & {\bf Seq3} \\ \hline
		Our approach & 80.6 \% & 82.1 \% & 81.2 \% \\
		CEM \cite{MilanRS14pami}  & 72.8 \% & 76.0 \%  & 76.8 \%\\
		DCO \cite{SR12} & 72.4 \% & 74.3 \% & 74.7 \% \\
		Baseline(Kalman Filter)  & 59.8 \% & 62.4 \% &  62.0\% \\
		\hline
	\end{tabular}
	\caption{
		\label{tabmotp}
		The MOTP in different test sequences.}
\end{table}

\begin{table}[!hbp]
	\center
	\begin{tabular}{c | c | c | c}
		\hline
		{\bf Tracking Algorithm} & {\bf Seq1}  & {\bf Seq2} & {\bf Seq3} \\ \hline
		Our approach & 63.8 \% & 65.1 \% & 64.6 \% \\
		CEM \cite{MilanRS14pami}  & 60.5 \% & 62.1 \%  & 61.6 \%\\
		DCO \cite{SR12} & 58.7 \% & 60.2 \% & 59.9 \% \\
		Baseline(Kalman Filter)  & 48.6 \% & 50.1 \% & 50.7 \% \\
		\hline
	\end{tabular}
	\caption{
		\label{tabmota}
		The MOTA in different test sequences.}
\end{table}

\begin{table}[!hbp]
	\center
	\begin{tabular}{c | c | c}
		\hline
		{\bf Tracking Algorithm} & {\bf MOTA}  & {\bf MOTP} \\ \hline
		Our approach & 64.5 \% & 81.3 \% \\
		CEM \cite{MilanRS14pami}  & 61.4 \% & 75.2 \% \\
		DCO \cite{SR12} & 59.6 \% & 73.8 \% \\
		Baseline(Kalman Filter)  & 49.8 \% & 61.4 \% \\
		\hline
	\end{tabular}
	\caption{
		\label{tab}
		Comparison of tracking performance with MOTA and MOTP.}
\end{table}

%The speed of our tracking method can be about four frames per second. However, we have to admit that our approach cannot be real time. Owing to employing DPM, our detection stage is fairly slow. A faster detector is expected to be designed in the future work.
%\begin{tabular}{c | c | c | c}
% & {\bf Benchmark} & {\bf Easy} & {\bf Moderate} & {\bf Hard}\\ \hline
%Car (Detection) & 83.63 \% & 85.74 \% & 76.71 \%\\
%Car (Orientation) & 34.00 \% & 35.45 \% & 31.89 \%\\
%\end{tabular}

\section{Conclusion and Future Work} \label{conclusion}
In summary,  a novel tracking-by-detection framework is proposed in this paper. Our approach captures rich  information about road junctions, such as vehicle shape  and motion priors. As a consequence, the proposed approach has higher efficiency than traditional tracking algorithms in crowded scenes.  Though our approach is  tested in road intersections, by applying pedestrian detectors the proposed method is also suitable for other crowded situations, such as supermarket and subway station.             The main contributions of this work are as follows. First, we exploit shape prior  in the  sequential detection model to tackle occlusions in crowded scene. Second, Traffic force is defined  to model group behavior in the traffic scene. With GBM, we can handle the influence of  neighboring vehicles and obtain  more precise localizations. The proposed framework is evaluated on real traffic videos and has shown its significant performance through intensive comparisons and analyses. 

However, the proposed tracking-by-detection framework still can be improved. A faster  vehicle detector particularly designed  for  the traffic scene is expected in the future work. Besides, we also plan to judge whether the vehicles violate the traffic rules  on the basis of this work in the future.

\ifCLASSOPTIONcaptionsoff
  \newpage
\fi

\bibliographystyle{IEEEtran}
\bibliography{IEEEabrv,reference}

\begin{IEEEbiographynophoto}{Yuan Yuan} (M'05-SM'09) is currently a full professor with the School of Computer Science and Center for OPTical IMagery Analysis and Learning (OPTIMAL), Northwestern Polytechnical University, Xi'an 710072, Shaanxi, P. R. China. She has authored or coauthored over 150 papers, including about 100 in reputable journals such as IEEE Transactions and Pattern Recognition, as well as conference papers in CVPR, BMVC, ICIP, and ICASSP. Her current research interests include visual information processing and image/video content analysis.
\end{IEEEbiographynophoto}

\begin{IEEEbiographynophoto}{Yuwei Lu} received the B.E. degree in Software Engineering from the Northwestern Polytechnical University, Xi'an 710072, Shaanxi, P. R. China, in 2015. He is currently pursuing the Ph.D. degree from the School of Computer Science and the Center for Optical Imagery Analysis and Learning (OPTIMAL), Northwestern Polytechnical University, Xi'an 710072, Shaanxi, P. R. China. His research interests include computer vision and pattern recognition.
\end{IEEEbiographynophoto}

\begin{IEEEbiographynophoto}{Qi Wang} (M'15-SM'15) received the B.E. degree in automation and Ph.D. degree in pattern recognition and intelligent system from the University of Science and Technology of China, Hefei, China, in 2005 and 2010 respectively. He is currently an Associate Professor with the School of Computer Science and the Center for Optical Imagery Analysis and Learning (OPTIMAL), Northwestern Polytechnical University, Xi'an 710072, Shaanxi, P. R. China. His research interests include computer vision and pattern recognition.
\end{IEEEbiographynophoto}

\end{document}